\documentclass[11pt]{article}

\usepackage{coling2020}
\usepackage{times}
\usepackage{url}
\usepackage{latexsym}
\usepackage{covington}
\usepackage{multirow}
\usepackage{rotating}
\usepackage[round, authoryear]{natbib}
\usepackage{linguex}
\usepackage{hyperref}
\usepackage{arydshln}

\usepackage{graphicx}
\usepackage{caption, subcaption}
\usepackage{pgfplots}
\pgfplotsset{width=7cm,compat=1.8}
\usepackage{wrapfig}

\usepackage{booktabs}

\usepackage{enumitem}

\usepackage{amsmath}
\usepackage{amssymb}

\def\bert{BERT}
\def\roberta{RoBERTa}
\def\albert{ALBERT}

\title{A Closer Look at Linguistic Knowledge in Masked Language Models: \\ The Case of Relative Clauses in American English}

\author{Marius Mosbach\thanks{~~~equal contribution.}~, \textbf{Stefania Degaetano-Ortlieb}\footnotemark[1]~, \textbf{Marie-Pauline Krielke}, \\ \textbf{Badr M. Abdullah}, \textbf{Dietrich Klakow} \\ 
         Department of Language Science and Technology, Saarland University, Germany \\ 
         \normalsize \texttt{\{mmosbach,babdullah,dklakow\}@lsv.uni-saarland.de} \\
         \normalsize \texttt{s.degaetano@mx.uni-saarland.de} \\
          \normalsize \texttt{mariepauline.krielke@uni-saarland.de} 
          }

\date{}

\begin{document}

\maketitle

\begin{abstract}

Transformer-based language models achieve high performance on various tasks, but we still lack understanding of the kind of linguistic knowledge they learn and rely on. We evaluate three models (\bert{}, \roberta{}, and \albert{}), testing their  grammatical and semantic knowledge by sentence-level probing, diagnostic cases, and masked prediction tasks. We focus on relative clauses (in American English) as a complex phenomenon needing contextual information and antecedent identification to be resolved. Based on a naturalistic dataset, probing shows that all three models indeed capture linguistic knowledge about grammaticality, achieving high performance. Evaluation on diagnostic cases and masked prediction tasks considering fine-grained linguistic knowledge, however, shows pronounced \textit{model-specific} weaknesses especially on semantic knowledge, strongly impacting  models' performance. Our results highlight the importance of (a) model comparison in evaluation task and (b) building up claims of model performance and the linguistic knowledge they capture beyond purely probing-based evaluations.

\end{abstract}

\section{Introduction}
\label{intro}
\blfootnote{
    %
    %
    %
    %
    %
    
    \hspace{-0.65cm}  
    This work is licensed under a Creative Commons 
    Attribution 4.0 International License.
    License details:
    \url{http://creativecommons.org/licenses/by/4.0/}.
}
Endeavors to better understand transformer-based masked language models (MLMs), such as BERT, are ever growing since their introduction in 2017 (cf.\ \citet{Rogers-etal2020} for an overview). While the BERTology movement has enhanced our knowledge on the reasons behind BERT's performance in various ways, still plenty remains unanswered. Less well studied and challenging are linguistic phenomena, where, besides contextual information, identification of an antecedent is needed, such as relative clauses (RCs). \citet{Kim-etal2019}, e.g., analyzed BERT's  comprehension of function words, showing how relativizers and prepositions are quite challenging for BERT. Similarly, \citet{WarstadtBowman2019} find RCs to be difficult for BERT in the CoLA acceptability tasks. 

In this paper, we focus on RCs in American English to further enhance our understanding of the grammatical and semantic knowledge captured by pre-trained MLMs, evaluating three models:\ BERT, RoBERTa, and ALBERT. For our analysis, we train probing classifiers, consider each models' performance on diagnostic cases, and test predictions in a masked language modeling task on selected semantic and grammatical constraints of RCs. RCs are clausal post-modifiers specifying a preceding noun phrase (antecedent) and are introduced by a relativizer (e.g., \textit{which}). Extensive corpus research \citep{Biber-etal1999} found that the overall most common relativizers are \textit{that}, \textit{which}, and \textit{who}. The relativizer occupies the subject or object position in a sentence (see examples~\Next[a] and~\Next[b]).  In subject RCs, the relativizer is obligatory \citep[1055]{HuddlestonPullum2002}, while in object position omission is licensed (e.g., \textit{zero} in example~(1-b)).

\ex. \label{ex:1}
\a. {\sl Children \textit{who} eat vegetables are likely to be healthy. (subject relativizer, relativizer is obligatory)}
\b. {\sl This is the dress [\textit{that/which/zero}] I brought yesterday. (object relativizer, omission possible)}

\noindent Relativizer choice depends on an interplay of different factors.\footnote{These factors include register (fiction, news, academic texts), restrictiveness (restrictive RCs add information about the head noun necessary for identification of the latter (see Example~\ref{ex:1}); non-restrictive RCs add information elaborating on a head noun which ``is assumed to be already known” \citet[602]{Biber-etal1999} and  are usually separated by a comma, e.g.\ \textit{My children, who I love, eat vegetables.}), animacy of the head noun (animate or inanimate antecedent), American (AE) vs.\ British English (BE), and definiteness of a pronominal antecedent (demonstrative vs.\ indefinite pronoun).} Among these factors, the animacy constraint \citep{Quirk1957} is near-categorical:\ for animate head nouns the relativizer \textit{who} (see Example~1) is strongly prioritized (especially over \textit{which}) \citep{DarcyTagliamonte2010}.

Our aims are (1) to better understand  whether sentence representations of pre-trained MLMs capture grammaticality in the context of RCs, (2) test the generalization abilities and weaknesses of probing classifiers with complex diagnostic cases, and (3) test prediction of antecedents and relativizers in a masked task considering also linguistic constraints. From a linguistic perspective, we ask whether MLMs correctly predict (a)  grammatically plausible relativizers given certain types of antecedents (animate, inanimate) and vice versa grammatically plausible antecedents given certain relativizers (\textit{who} vs.\ \textit{which/that)}, and (2) semantically plausible antecedents given certain relativizers  considering the degree of specificity of predicted antecedents in comparison to  target antecedents (e.g.\ \textit{boys} as a more specific option than \textit{children} in Example~\ref{ex:1}). 
Moreover, we are interested in how these findings agree with  probing results and investigate model specific behavior, evaluating and comparing the recent pre-trained MLMs:~BERT, RoBERTa, and ALBERT. This is to our  knowledge the first attempt comparing and analyzing performance of different transformer-based MLMs in such detail, investigating grammatical and semantic knowledge beyond probing.

Our main contributions are the following:\ (1) the creation of a naturalistic dataset for probing, (2) a detailed model comparison of three recent pre-trained MLMs, and (3) fine-grained linguistic analysis on  grammatical and semantic knowledge. 
Overall, we find that all three MLMs show good performance on the probing task. Further evaluation, however, reveals model-specific  issues with wrong agreement (where \roberta{} is strongest) and distance between antecedent-relativizer and relativizer-RC verb (on which \bert{} and \albert{} are better). Considering linguistic knowledge, all models perform better on grammatical rather than semantic knowledge. Out of the relativizers, \textit{which} is hardest to predict. Considering model-specific differences, \bert{} outperforms the others in predicting the actual targets, while \roberta{} captures best grammatical and semantic knowledge. \albert{} performs worst overall.

\section{Background}

\subsection{Models}

BERT \citep{devlin-etal-2019-bert} is a transformer-based \citep{NIPS2017_7181} bidirectional network trained on masked language modeling and next-sentence-prediction. The extent to which BERT captures linguistic knowledge is widely studied in previous works (see §\ref{sec:related_work}). RoBERTa \citep{liu2019roberta} differs from BERT in four important aspects:\ trained on more data, no next-sentence-prediction objective, achieves lower perplexity on the training data, while using  larger vocabulary. Given RoBERTa's superior performance over BERT on the GLUE benchmark \citep{wang-etal-2018-glue}, RoBERTa has replaced BERT as the model of choice in several recent studies investigating MLMs \citep{pruksachatkun-etal-2020-intermediate, talmor2020teaching}. Nevertheless, RoBERTa's linguistic properties and how they differ from BERT's remain relatively unexplored. ALBERT \citep{Lan2020ALBERT:} is a recently proposed alternative to BERT and RoBERTa. It uses weight-sharing across all hidden layers --- effectively applying the same non-linear transformation at every layer --- and factorizes the embedding matrix into two separate matrices, resulting in significantly fewer model parameters compared to BERT and RoBERTa. ALBERT was shown to outperform BERT and RoBERTa when fine-tuned on several English NLP downstream tasks \citep{Lan2020ALBERT:}. However, to our knowledge, no previous work has systematically evaluated the linguistic knowledge of ALBERT. For all models, we consider the base variants: \textit{BERT-base-cased}, \textit{RoBERTa-base}, \textit{ALBERT-base-v1} with $110$M, $125$M, and $12$M parameters, respectively.

\subsection{Related Work}
\label{sec:related_work}

Related work has investigated grammaticality of unidirectional language models using \textit{Minimal Pair Evaluation} \citep{marvin-linzen-2018-targeted,wilcox-etal-2019-structural, warstadt2019blimp,Hu2020ACL,hu-etal-2020-systematic}. Out of work on MLM prediction-based evaluation  (e.g., \citet{Goldberg2019, ettinger-2020,petroni-etal-2019-language,jiang2019can,kassner-schutze-2020-negated}), only \citet{Goldberg2019} has so far evaluated MLM predictions in the context of grammaticality. Our study adds to this line of work, focusing on RCs 
combining MLM prediction based evaluation with probing. 

Work related to evaluating sentence embeddings considered prediction of sentence length, word content, and word order \citep{adi2016fine}. \citet{conneau-etal-2018-cram} investigated an even broader range of linguistic properties.  
Extensive work has been done on probing pre-trained 
MLMs, especially BERT, for syntactic and semantic knowledge (see e.g., \citet{tenney2018what, liu-etal-2019-linguistic} and \citet{Rogers-etal2020} for a more comprehensive overview). 

Most similar to our work is \citet{warstadt-etal-2019-investigating}, focusing on comparing evaluation methods, including probing and MLM evaluation to assess how models encode linguistic features. 
We contribute to this strand of research by building datasets from naturalistic (rather than artificial as in \citet{warstadt-etal-2019-investigating}) data and comparing three transformer models: BERT, RoBERTa, and ALBERT. Our focus is on RCs as challenging sentence types for pre-trained language models as shown by, e.g., 
\citet{Kim-etal2019} (showing  relativizers to be quite challenging for BERT) and \citet{WarstadtBowman2019} (finding RCs difficult for BERT in the COLA probing tasks).

\section{Probing MLMs Representations for Grammatical Knowledge of RCs}

We train supervised \textit{probing} classifiers (here:\ acceptability classifiers) to assess the linguistic knowledge contained in a model's representations, focusing on grammaticality. A model's grammaticality awareness should be reflected in the hidden representations produced by the model for that particular sentence. Hence, by training a classifier on top of these representations, we should be able to discriminate between grammatical and ungrammatical sentences based on their sentence embeddings.\footnote{Here, we assume that grammatical and ungrammatical sentences are \textit{linearly separable} in the embedding space.} 
Moreover, besides capturing knowledge on grammaticality, we also examine whether this knowledge becomes more or less separable in the representation, considering representations produced by different layers of a model.

\subsection{Dataset Construction}
\label{sec:probing-dataset}

\begin{table}[t]
\resizebox{.99\textwidth}{!}{
\begin{tabular}{@{}ccccl@{}}
\toprule
\textbf{Animate} & \textbf{Restrictive} & \textbf{SubjRC} & \textbf{Modification}              & \multicolumn{1}{c}{\textbf{Sentence}}                                                                  \\ \midrule
1       & 1          & 1          & no modification                      & Katrina Haus was a woman \textbf{who} sought to attract stares, not turn them away.                    \\ 
1       & 1          & 1          & who $\rightarrow$ which  & *Katrina Haus was a woman \textbf{which} sought to attract stares, not turn them away.                  \\ \midrule
0       & 1          & 0          & no modification  & She pulls out a course catalog, various forms, and a letter \textbf{which} she hands to Kevin.                   \\
0       & 1          & 0          & which $\rightarrow$ that & She pulls out a course catalog, various forms, and a letter \textbf{that} she hands to Kevin.                      \\ \midrule
0       & 1          & 0          & no modification & Never permit your muzzle to cover anything \textbf{which} you are unwilling to shoot.                  \\
0       & 1          & 0          & relativizer omission & Never permit your muzzle to cover anything you are unwilling to shoot.                  \\ \midrule
0       & 0          & 0          & no modification & I never saw a penny in royalties, \textbf{which} was all right with me.                        \\
0       & 0          & 0          & which $\rightarrow$ who & *I never saw a penny in royalties, \textbf{who} was all right with me.                        \\
\bottomrule

\end{tabular}
}
\caption{Examples from the dataset (minimal pairs). The relativizer
is shown in bold. Modifying a sentence does not always result in an ungrammatical sentence. For example, when Restrictive=1 and SubjRC=0, relativizer omission yields a grammatical sentence. 
}
\label{table:examples_from_dataset}
\end{table}

To probe pre-trained MLMs' performance on sentences containing RCs, we construct a controlled set of 48,060 sentences  and their acceptability labels using an automated procedure. Our probing dataset is a subset of naturally occurring sentences extracted from the fiction portion of the COCA corpus \citep{davies}. First, we extract all sentences containing only one pronoun from the set \{\textit{who, whom, whose, which, that}\}. We then parse the sentences using SpaCy dependency parser and keep only those sentences where the pronoun constitutes a relativizer (identified by the tag \textsc{relcl}). From the parse tree, we also determine whether the relativizer fills the subject or the object position in the RC. The automatic selection procedure is illustrated in Fig.\ \ref{fig:appendix_data_creation} in  §\ref{sec:prob_data}. 

On top of grammatical sentences from the corpus, we manipulate the data obtaining a set with unacceptable ones. To this end, we populate three Boolean meta-data variables for each grammatical sample in our data using a set of hand-crafted linguistic rules: \textsc{animate}, \textsc{restrictive}, and \textsc{subjrc}. Based on the values of these three meta-data variables, a set of modifications is applied to convert a grammatical sentence into an ungrammatical one. The procedure of creating the dataset is explained in detail in  §\ref{sec:prob_data}. Our final dataset consists of 42.7k and 5.3k samples for training and evaluation, respectively. Both splits are balanced, i.e. accuracy of the majority baseline is $50\%$. Table \ref{table:examples_from_dataset} presents a set of minimal pair examples that are generated using our procedure.

\subsection{Experimental Setup}
\label{sec:setup}

We train logistic regression acceptability classifiers on sentence embeddings obtained from the hidden layers of a pre-trained model. We compute sentence embeddings of BERT, RoBERTa, and ALBERT and two non-contextualized baselines:\ GloVe embeddings \citep{pennington-etal-2014-glove} trained on English Wikipedia and the Gigaword corpus and fasttext embeddings \citep{bojanowski2016enriching} trained on English Wikipedia and News data \citep{mikolov2018advances}. As an additional baseline we use a rule based classifier that simply classifies sentences containing a relativizer (who, which, that) as grammatical and as ungrammatical otherwise.
\begin{wrapfigure}{r}{0.40\textwidth}
    \vspace{-15pt}
  \begin{center}
    \includegraphics[trim=30 30 10 10, width=.40\textwidth]{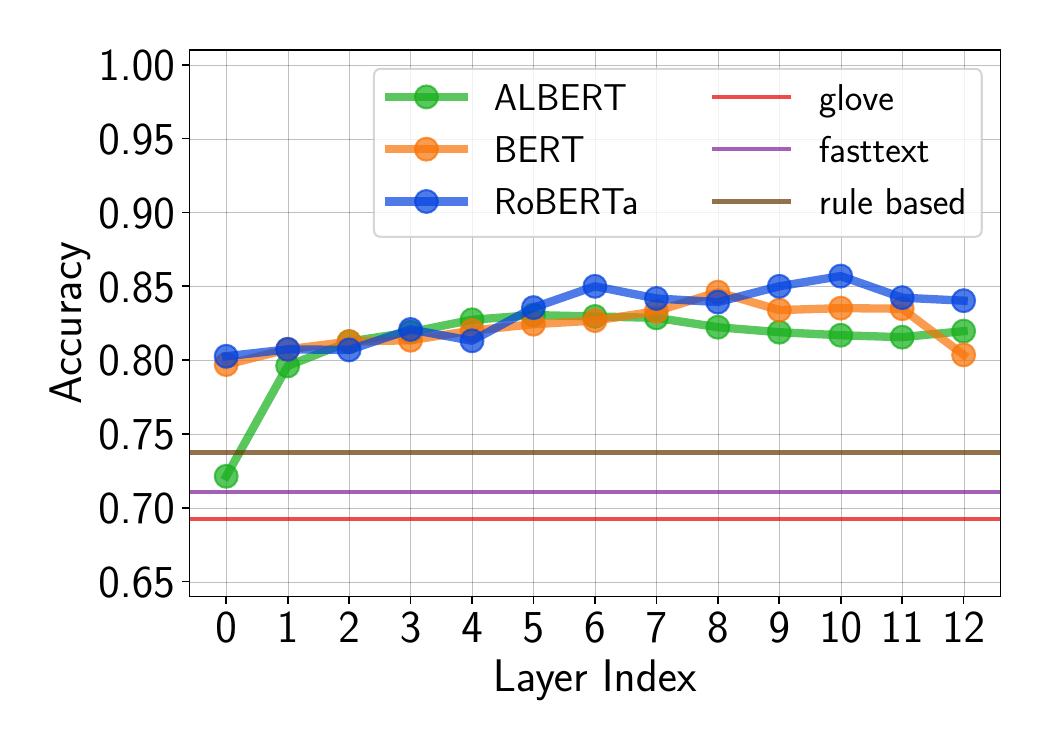}
  \end{center}
  \vspace{-10pt}
  \caption{Layer-wise probing accuracy on the test set for pre-trained transformer and baseline models using mean-pooling.}
  \label{fig:layer-wise-accuracy}
  \vspace{18pt}
\end{wrapfigure}
\indent Input sentences are pre-processed by adding two special tokens, [CLS] and [SEP], at the beginning and end of each input sentence, respectively.\footnote{To be precise, for RoBERTa the special tokens are $<$s$>$ and $<$/s$>$. For GloVe and fasttext embeddings as well as the rule base classifier, we tokenize on the word level without using special tokens.} To construct sentence embeddings, we apply two different pooling strategies:\ CLS- and mean-pooling.\footnote{CLS-pooling simply returns the vector representation of the first token of the input sentence, i.e. the [CLS] token. Mean-pooling computes a sentence embedding by taking the mean over all (sub-word) token representations of the input sentence.} We obtain sentence embeddings from all hidden layers of the MLM models, including the non-contextualized embedding layer (layer 0). We treat accuracy on the acceptability classification task as a proxy for the linguistic knowledge encoded in a model's sentence embeddings, which we evaluate on a held-out test set. We use the huggingface transformers \citep{wolf2019huggingface} and flair \citep{akbik2018coling} libraries as well as Scikit-learn \citep{scikit-learn} for obtaining embeddings and training the logistic regression classifiers. Code to reproduce our probing dataset and results is available online: \url{https://github.com/uds-lsv/rc-probing}.

\subsection{Probing Results and Discussion}
\label{sec:Mperf}

Fig.~\ref{fig:layer-wise-accuracy} shows the layer-wise probing accuracy for all models when using mean-pooling. A side-by-side comparison with CLS-pooling is shown in Fig.~\ref{fig:appendix:probing_results} in §\ref{appprob}.
Mean-pooling leads to significantly higher probing accuracies for all models, suggesting a sub-optimal encoding of sentence-level information in the CLS token representation. Hence, we stick to mean-pooling in the following. We find that the rule base classifier is a surprisingly strong baseline, outperforming both the GloVe and fasttext baselines. Moreover, Fig.~\ref{fig:layer-wise-accuracy} shows that all three transformer-based models improve significantly over the baselines for almost all layers. The only exception is layer 0 of ALBERT, which performs similarly to the fasttext baseline and worse than the rule based classifier. We attribute this finding to the embedding factorization of ALBERT. At lower (contextualized) layers (1--5),  probing accuracies of BERT, RoBERTa, and ALBERT are almost identical. For higher layers, both BERT and RoBERTa improve over ALBERT, with ALBERT's probing accuracy remaining roughly constant. We note that ALBERT has significantly less parameters than BERT and RoBERTa (12M vs.\ 110M and 125M), which might explain the lower probing accuracy. We provide a more detailed discussion, investigating the role of the number of parameters, in §\ref{sec:alberts}. Overall, the fact that probing accuracy is above $80\%$ for almost all layers suggests a reasonable encoding of sentence-level linguistic knowledge relevant for grammaticality classification in all pre-trained models. Notably, linear separability with respect to grammaticality emerges very early in the sentence embeddings of all three models.

To get a better understanding of the accuracy achieved by each of the models, we select the best classifier 
according to the results in Fig.~\ref{fig:layer-wise-accuracy} and report test accuracy grouped by modification in Table \ref{tab:probing-results}. For comparison, we additionally report accuracies of the non-contextualized baselines (layer 0) of each model in parentheses. The results show that while contextualization leads to higher probing accuracy overall, it is especially important for the \texttt{which $\rightarrow$ who} and \texttt{which $\rightarrow$ that} 
samples. From a linguistic viewpoint this is surprising: e.g. replacing \textit{which} by \textit{who} clearly makes a sentence ungrammatical and is typically easy to detect for humans. When looking at the training data, however, this observation is not surprising at all. The vast majority of sentences (15k samples) belong to the \texttt{no modification} group contain \textit{who} as a relativizer. All of these sentences are grammatical (cf. Table~\ref{tab:stats_mods}). On the other hand, \texttt{which $\rightarrow$ who} contains only 2.5k samples, all of them are ungrammatical. Hence, our results in Table \ref{tab:probing-results} reveal that the non-contextualized baselines might learn a simple heuristic, classifying all sentences that contain \textit{who} as a relativizer as grammatical. The results of the rule based baseline classifier give further evidence for this interpretation, showing that a simple classifier which bases its predictions only on the existence of a relativizer has a surprisingly high accuracy (comparable to the non-conextualized baselines) on our dataset. Interestingly, BERT and RoBERTa seem to be especially susceptible to this short-cut learning as shown by the results of the non-contextual baselines for BERT and RoBERTa on the \texttt{relativizer omission} and \texttt{which $\rightarrow$ who} modifications.

\subsection{Diagnostics}\label{sec:diag} 

\begin{table}[t]
    \centering \small
        \begin{tabular}{lccccccc}
        \toprule
              \textbf{Modification} &  \textbf{GloVE} &  \textbf{fasttext} &  \textbf{rule based} & \textbf{BERT} &  \textbf{RoBERTa} &  \textbf{ALBERT}  \\ \midrule
              no modification & $67.3$ & $69.2$  & $100$ & $83.7$ $(78.4)$ & $85.3$ $(79.7)$ & $83.6$ $(71.5)$  \\
              relativizer omission  & $77.2$ & $85.3$ & $97.6$ & $96.0$ $(98.3)$ & $95.1$ $(98.2)$ & $96.4$ $(94.8)$  \\
              who $\rightarrow$ which & $88.0$ & $81.7$ & $0.00$ & $84.9$ $(88.4)$ & $87.3$ $(89.6)$ & $78.8$ $(68.3)$  \\
              which $\rightarrow$  who & $18.2$ & $21.3$ & $0.00$ & $47.9$ $(1.73)$ & $50.0$ $(0.02)$ & $44.2$ $(18.7)$  \\
              which $\rightarrow$  that & $12.8$ & $8.0$ & $100$ & $80.0$ $(52.0)$ & $77.6$ $(44.8)$ & $72.0$ $(22.4)$  \\ \midrule
              total &  $69.2$ &  $71.0$ & $73.7$ &  $84.6$ $(79.7)$ &  $85.5$ $(80.1)$ &  $83.0$ $(72.1)$ \\ 
        \bottomrule     
        \end{tabular}
    \caption{Test accuracy (in $\%$) grouped by modification type (cf. Table~\ref{tab:stats_mods} for statistics). For BERT, RoBERTa, and ALBERT we select the best model according to the probing results shown in Fig.~\ref{fig:layer-wise-accuracy}. Numbers in parenthesis show the accuracy of the non-contextualized baseline (layer 0) for each model. 
    }
    \label{tab:probing-results}
\end{table}

To investigate generalization abilities and model specific behavior of the best probing classifiers, we evaluate them on a diagnostics dataset containing sentences with the following properties:\ (1) adjacent antecedent and relativizer (see example \Next[a], grammatical), (2) longer distance between  antecedent and  relativizer (see~ example \Next[b], grammatical), (3) longer distance between relativizer and RC verb (see example \Next[c], grammatical), (4) wrong agreement between adjacent antecedent and relativizer (see example \Next[d], ungrammatical), and (5) intervening agreement attractors leading to wrong agreement (see~\Next[e], ungrammatical). To create the dataset, we manually select four sentences and manipulate these according to each of the above mentioned cases. Three sentences have nominal antecedents 
and one has a clausal antecedent. 
For the grammatical manipulations (case 0--2), we additionally test between restrictive/non-restrictive variants of each sentence. Overall, we test 
a total of 32 sentences\footnote{We opted for a small set for high control over factors influencing a choice.} and evaluate the models' confidence based on the un-normalized log probabilities (logit), where predictions $> 0$ result in classifying sentences as grammatical. 
In the manipulations according to case 2, besides considering nominal vs.\ clausal antecedent, we look at the number of words of intervening phrases (length of 3--7 intervening words).\footnote{Note that for case 3, distance between relativizer and verb is always three words, so length is not a factor to be evaluated.} 

\ex. \label{ex:diagnostics}
\a. {\sl We just heard \textbf{a debate}        \textbf{which} was about the differences in wage rates [...].
    (case 0:\ adjacent antecedent and relativizer)}
\b. {\sl [...] \textbf{a debate} on one of the most famous television channels       \textbf{which} was about [...].
    (case 1:\ distance antecedent to relativizer)}
\c. {\sl [...] a debate \textbf{which} in many regards \textbf{was} an important     one about  [...].
    (case 2:\ distance relativizer to verb)}
\d. {\sl [...] \textbf{a debate who} was about [...].
    (case 3:\ wrong agreement antecedent-relativizer)}
\e. {\sl [...] a debate \textbf{by DeGeneres} who was about [...].
    (case 4:\ agreement attractor)}

\begin{table}[t]
    \centering \small
    \begin{tabular}{clcccccc}
    \toprule
         \multirow{2}{*}{\textbf{Case}} & \multirow{2}{*}{\textbf{Factor}} & \multicolumn{2}{c}{\textbf{BERT}} & \multicolumn{2}{c}{\textbf{ALBERT}} & \multicolumn{2}{c}{\textbf{RoBERTa}} \\
         \cmidrule{3-8}
        & & restrictive & non-restrictive & restrictive & non-restrictive & restrictive & non-restrictive \\
        \midrule
        \multirow{2}{*}{1 (+)} & nominal &~~~$2.74$ &$1.62$&~~~$2.35$&$1.75$&$-0.79$&$-0.85$\\
        & clausal &$-0.39$ &$0.89$&$-1.18$&$0.57$&$-0.85$&$-0.57$\\
        \midrule
        \multirow{4}{*}{2 (+)} & nominal &~~~$2.34$ &$1.74$&~~~$1.30$&$1.72$&$-1.03$&$-0.87$\\
        & clausal &$-0.39$ &$0.89$&$-1.18$&$0.57$&$-0.85$&$-0.57$\\
        \cmidrule{2-8}
        & $3$--$4$ words & ~~~$2.35$ & $2.52$ & ~~~$1.30$ & $2.55$ & $-1.03$ & $-0.55$ \\
        & $>$ 4 words & $-0.69$ & $0.54$ & $-1.15$ & $0.31$ & $-1.42$ & $-1.04$ \\
        \midrule
        \multirow{2}{*}{3 (+)} & nominal & ~~~$1.99$ & $1.97$ & ~~~$2.04$ & $1.78$ & $-1.06$ & $-0.80$ \\
        & clausal & $-0.84$ & $0.43$ & $-1.54$ & $0.17$ & $-0.89$ & $-0.35$ \\
        \midrule
        \multirow{2}{*}{4 (*)} & nominal & $-0.47$ & -- & $-0.40$ & -- & $-1.46$ & -- \\
        & clausal & $-1.38$ & -- & $-1.75$ & -- & $-1.24$ & -- \\
        \midrule
        \multirow{2}{*}{5 (*)} & nominal & ~~~$0.56$ & -- & ~~~$0.08$ & -- & $-1.24$ & -- \\
        & clausal & $-1.38$ & -- & $-1.75$ & -- & $-1.24$ & -- \\
        \bottomrule
    \end{tabular}
    \caption{Prediction confidence in mean logit ($>0$:\ grammatical, $< 0$:\ ungrammatical). Sentences in case 0 to 2 should be predicted as grammatical (+) and case 3 and 4 as ungrammatical (*).
    }
    \label{tbl:dia}
\end{table}

\textbf{Case 1.} For restrictive and non-restrictive variants and nominal antecedents, \bert{} and \albert{} correctly predict with relatively high confidence the sentences to be grammatical (restrictive:\ $2.7$,   non-restrictive:\ $1.6$). For the clausal antecedent, they fail in the restrictive variant ($-0.39$ and $-1.18$, respectively), while they can deal with  non-restrictive RCs ($0.89$ and $0.57$),  the comma possibly being a strong indicator of non-restrictiveness.  \roberta{} always wrongly predicts ungrammaticality. 

\textbf{Case 2.} With intervening phrases between antecedent and relativizer (see example (2-b)), \bert{} and \albert{} again deal well in both restrictive and non-restrictive RCs for  nominal antecedents, but fail for restrictive RCs in the clausal case. \roberta{} again fails to predict the grammatical class.  Considering the length of the intervening phrase, for restrictive RCs, BERT deals with intervening phrases  
with high confidence ($2.35$), provided 
they are relatively short (3--4 words). For longer distance, \bert{} is less confident ($-0.69$) and chooses the (wrong) ungrammatical class.  
ALBERT acts similarly, but fails with higher confidence ($-1.15$). RoBERTa always fails without a clear pattern. For the non-restrictive RCs, both BERT and ALBERT correctly predict the grammatical class, even though distance affects 
their confidence (the longer the less confident:\ $2.52$ for shorter, $0.54$ for longer phrases). RoBERTa again predicts the ungrammatical class for all sentences, but is especially confident for greater distance between antecedent and relativizer ($-0.55$ for shorter, $-1.04$ for longer phrases).

\textbf{Case 3.} With restrictive RCs, RoBERTa has problems  with intervening phrases between relativizer and  RC verb, again always predicting the ungrammatical class. BERT predicts this case correctly, but fails at sentences with clausal antecedent, even though with lower confidence than \albert{} ($-0.84$ vs.\ $-1.54$).  In non-restrictive RCs, BERT and \albert{} obtain perfect accuracy for both nominal and clausal antecedents, being quite confident in their predictions, while  RoBERTa still fails with sometimes very high confidence.\footnote{ BERT and \albert{}  seem to rely on the comma as an indicator of grammaticality for non-restrictive RCs (cases 0--2).}

\textbf{Case 4.} With wrong agreement and adjacent nominal antecedent  and relativizer, RoBERTa is quite confident of the ungrammaticality of the sentences ($-1.46$), while BERT and \albert{} are confident in the clausal antecedent case ($-1.38$ and $-1.75$, respectively), but much less confident or even wrong with nominal antecedents.

\textbf{Case 5.} With intervening agreement attractors and wrong agreement, BERT  gets confused due to the attractors (see~ example \Last[e], where \textit{DeGeneres} is considered the antecedent instead of \textit{debate}) and is only confident in the clausal antecedent case ($-1.38$). \roberta{}, instead, is very confident in recognizing the wrong agreement ($-1.24$), while \albert{} gets confused as well, even though less and with much less confidence than BERT ($0.08$).  

In summary, \roberta{} is quite confident in wrong agreement cases. 
BERT and \albert{}     deal much better than \roberta{} with longer distances between antecedent and relativizer. Also, BERT and \albert{}   learn to recognize non-restrictive RCs quite well and can deal with phrases between relativizer and RC verb. Thus, even though the models achieve very high probing accuracy overall (see §~\ref{sec:Mperf}), evaluating on more complex cases reveals that each model seems to rely on different kinds of information, strongly affecting the generalization abilities of the probing classifiers.  While we are aware of the diagnostics set's very limited extend, hindering generalizable conclusions, the controlled diagnostics evaluation gives an indication of possible differences underlying prediction choices across models.

\section{Analyzing MLMs Predictions for RC Awareness}

\subsection{Motivation}

For a more comprehensive picture of the differences between models, we perform a masked language modeling evaluation \citep{Goldberg2019}, looking at the models' predictions of relativizers as well as antecedents. Besides grammaticality, we also test whether the models capture semantic knowledge. 

\subsection{Analyzing Grammatical and  Semantic Knowledge}

We extract sentences containing restrictive\footnote{We deliberately exclude non-restrictive RCs due to the indicative comma facilitating RCs recognition (see §~\ref{sec:diag}).} object or subject RCs with one of the three relativizers 
from the registers magazines and academic\footnote{We exclude fiction as it was used for the probing experiments, and spoken, due to possible high noise in spoken data.} from the COCA corpus \citep{davies}. All sentences are formatted as described in §~\ref{sec:setup}, e.g. [CLS] The woman [MASK] studies linguistics. [SEP]. The size of each set of RC type depends on the frequency of available sentences in the COCA corpus and therefore varies between types from 20 to 50 sentences.

\subsubsection{Relativizer Prediction}

\begin{table}[t]
    \centering
    \begin{subtable}{.49\textwidth}{
        \resizebox{.95\textwidth}{!}{%
        \begin{tabular}{llrrrrrr}
            \toprule
            &&\multicolumn{3}{c}{\textbf{objRC}}&\multicolumn{3}{c}{\textbf{subjRC}}\\	
            \midrule
            &	&who	&which	&that&	who	&which&	that\\
            \midrule
            \multirow{3}{*}{\begin{sideways}{\textbf{MP@1}}\end{sideways}}& BERT & $\textbf{0.80}$ &	$0.29$ &	$0.89$ &	$\underline{\textbf{0.98}}$ &	$0.02$ &	$\textbf{0.92}$ \\
            &RoBERTa & $\textbf{0.80}$ &	$0.24$ & \underline{$\textbf{0.96}$} & $0.92$ & $0.04$ &	$\textbf{0.92}$ \\
            &ALBERT& $0.32$ & $\textbf{0.43}$ & $0.86$ &	\underline{$0.96$} & $\textbf{0.16}$ &	$0.76$ \\
            \midrule
            \multirow{3}{*}{\begin{sideways}{\textbf{MTR}}\end{sideways}}& BERT& $\textbf{1.23}$ &	$1.83$ &	$1.11$ &	\underline{$\textbf{1.00}$} &	$2.46$ &	$\textbf{1.10}$ \\
            & RoBERTa & $\textbf{1.23}$ &	$1.95$ &	\underline{$\textbf{1.04}$} &	$1.08$ &	$2.18$ &	$1.22$ \\
            & ALBERT & $2.29$ &	$\textbf{1.73}$ &	$1.39$ &	\underline{$1.02$} &	$\textbf{1.18}$ &	$1.37$ \\
            \midrule
            \multirow{3}{*}{\begin{sideways}{\textbf{MNE}}\end{sideways}}& BERT& $0.09$ &	$0.08$ &	\underline{$\textbf{0.05}$} &	\underline{$0.05$} &	$0.07$ &	$0.06$ \\
            &RoBERTa& $\textbf{0.08}$&	$\textbf{0.06}$&	$\textbf{0.05}$&	\underline{$\textbf{0.04}$}&	$\textbf{0.05}$&	\underline{$\textbf{0.04}$}\\
            &ALBERT& $0.19$ &	$0.21$	& $0.18$&	\underline{$0.12$} &	$0.13$&	$0.17$\\
            \bottomrule 
        \end{tabular}
        }%
    }
    \caption{Metrical evaluation (MP@1:\ mean precision at 1,\\ MTR:\ mean target rank, MNE:\ mean normalized entropy)}
    \label{tab:relsprecis}
    \end{subtable}%
    ~    
    \begin{subtable}{.49\textwidth}{
        \resizebox{.95\textwidth}{!}{%
        \begin{tabular}{llrrrrrr}
            \toprule
            &&\multicolumn{3}{c}{\textbf{objRC}}&\multicolumn{3}{c}{\textbf{subjRC}}\\	
            \midrule
            &	&who	&which	&that&	who	&which&	that\\
            \midrule
            \multirow{3}{*}{\begin{sideways}{\textbf{AN}}\end{sideways}}&BERT & $\textbf{0.87}$ &	\underline{$\textbf{1.00}$} &	$0.93$ &	\underline{$\textbf{1.00}$} &	$\textbf{0.94}$ &	$\textbf{0.96}$\\
            &RoBERTa& $0.81$ &	\underline{$\textbf{1.00}$} &	$\textbf{0.96}$&	$0.94$ &	$0.90$ &	$0.94$\\
            &ALBERT& $0.71$ &	\underline{$\textbf{1.00}$} &	$0.93$ &	\underline{$\textbf{1.00}$} &	$\textbf{0.94}$ &	$0.92$ \\
            \midrule
            \multirow{3}{*}{\begin{sideways}{\textbf{PL}}\end{sideways}}&BERT&0.81&	$\textbf{0.98}$&	0.93&	\underline{$\textbf{1.00}$}&	0.92&	0.96\\
            &RoBERTa& $\textbf{0.97}$&	\underline{$\textbf{0.98}$}&	$\textbf{0.96}$& \underline{0.98}&	\underline{$\textbf{0.98}$}&	\underline{$\textbf{0.98}$}\\
            &\albert{}&0.68&	0.95&	0.89&	\underline{0.98}&	0.96&	0.92\\
            \midrule
            \multirow{3}{*}{\begin{sideways}{\textbf{GR}}\end{sideways}}&BERT&0.94&		\underline{$\textbf{1.00}$}&	0.93&		\underline{$\textbf{1.00}$}&	$\textbf{0.98}$&		\underline{$\textbf{1.00}$}\\
            &RoBERTa&$\textbf{0.97}$&		\underline{$\textbf{1.00}$}&	$\textbf{0.96}$&		\underline{$\textbf{1.00}$}&	$\textbf{0.98}$&		\underline{$\textbf{1.00}$}\\
            &\albert{}&0.68&	0.95&	0.89&		\underline{0.98}&		\underline{$\textbf{0.98}$}&	0.96\\
            \bottomrule 
            \end{tabular}
        }%
    }
    \caption{Semantic (AN:\ animacy, PL:\ plausibility) and \\ grammaticality (GR) evaluation.}
    \label{tab:relsqualitative}
    \end{subtable}
    \vspace{25pt}    
    \caption{\textit{Relativizer} prediction:\ Quantitative (a) and qualitative (b) evaluation. \textbf{Bold}:\ best result for each metric and category across models.\ \underline{Underline}:\ best result for each model per metric across categories.
    }
    \label{tab:relativizer}
\end{table}

We test all three models by masking the relativizer, considering the following metrics for evaluation:\ (1) mean precision at 1 (MP@1) (model's precision of target prediction at  first position), (2) mean target rank (MTR), and (3) normalized mean entropy (NME) (uncertainty of the model's prediction). For MP@1, all three models generally perform well across RCs apart from  \textit{which} (Table~\ref{tab:relsprecis}). BERT and RoBERTa  show similar performance, while \albert{} diverges slightly, with fairly weak predictions of \textit{who} in object RCs and comparatively accurate predictions of \textit{which} in both object and subject RCs. 
MTR is negatively correlated with MP@1 (the larger the divergence from 1, the lower the precision of the prediction). NME reflects the two other measures:\ BERT and RoBERTa are quite confident about their predominantly successful predictions, while \albert{} shows a higher uncertainty about overall much weaker predictions. 

Next, we manually evaluate\footnote{Evaluations were done by two linguistic experts.}
the actual predictions according to 
three criteria:\ \textit{animacy} (agreement between antecedent and relativizer), 
\textit{plausibility} (semantically plausible sentence) and \textit{grammaticality} (Table~\ref{tab:relsqualitative}). 
Results show that 
the actual felicity of the predictions is much stronger than  MP@1 would suggest, since non-target predictions are not necessarily infelicitous. This is especially true for \textit{which}, due to high interchangeability with \textit{that}. While both relativizers are synonymous, \textit{which} is primarily used in non-restrictive RCs. Since the sample sentences are restrictive, the models seem to predict the most frequent relativizer in restrictive RCs, which is \textit{that} (see~ example \ref{ex:thatwhich}). 

\ex. \label{ex:thatwhich}{\sl The action [MASK] it contemplates is command. (all models) (object RC, target=\ \textit{which}, prediction= that)}

\noindent Animacy is predicted reliably by BERT and RoBERTa, while \albert{} seems to have issues with \textit{who} object predictions. This is due to \albert{}’s preference to predict \textit{that} instead of \textit{who}, especially in fuzzy cases of general nouns describing humans (\textit{person, people, family, etc.}). Sometimes animacy is predicted correctly, however, case marking is infelicitous; the model seems to only take into account the subject and verb of the clause ignoring  possible clausal complements (see example \ref{ex:m4}).

\ex. \label{ex:m0}{\sl He spared no expense in moving, reassembling, and restoring buildings of people [MASK] he felt were the backbone of our nation. (\albert{}) (object RC, target=\ \textit{who}, prediction=\textit{that})}

\ex. \label{ex:m4}{\sl Jennifer had been having an online affair with a person [MASK] she believed was a man named Christopher. (\albert{}) (object RC, target=\ \textit{who}, prediction= \textit{whom})}

\noindent Accuracy for plausibility is very high, too, with \albert{} making the least plausible predictions. Note that plausibility entails grammaticality. Example~\ref{ex:m4} has animacy agreement, but is neither grammatical nor plausible. Grammaticality in contrast does not entail plausibility (see example \ref{ex:m5}). Here, BERT predicts the most frequent collocate of \textit{whirl} according to COCA \citep{davies}.

\ex. \label{ex:m5}{\sl This is something else, for I saw a whirl [MASK] I knew was a large bass. (BERT)  (object RC, target=\ \textit{which}, prediction= \textit{wind})}

\noindent Of all criteria, predictions are most accurate for grammaticality. Even \albert{} reaches a precision of 90\% on average.
Finally, we consider the ratio of (any) relativizers predicted vs.\ other word class, indicating how well the models recognize a syntagmatic environment typical for a RC to occur. RoBERTa, again, is most successful here (95\% of predictions are relativizers), followed by BERT (94\%) and \albert{} (90\%).

In summary, BERT and \roberta{} perform equally well at target prediction, while \roberta{} is most successful  qualitatively  (grammaticality,  plausibility, and relativizer prediction). Overall, our qualitative analysis  shows that all models largely make grammatical, plausible and animacy-conform predictions.

\begin{table}[t]
    \centering
    \begin{subtable}{.49\textwidth}{
        \resizebox{.95\textwidth}{!}{%
        \begin{tabular}{llrrrrrr}
            \toprule
            &&\multicolumn{3}{c}{\textbf{objRC}}&\multicolumn{3}{c}{\textbf{subjRC}}\\	
            \midrule
            &	&who	&which	&that&	who	&which&	that\\
            \midrule
            \multirow{3}{*}{\begin{sideways}{ \textbf{MP@1}}\end{sideways}}&\bert{}&\underline{$\textbf{0.38}$}&$\textbf{0.22}$&\underline{$\textbf{0.38}$}&$\textbf{0.27}$&$\textbf{0.31}$&$\textbf{0.29}$\\
            &\roberta{}&\underline{0.31}&0.08&0.29&0.22&0.17&0.14\\
            &\albert{}&0.28&0.14&\underline{0.29}&0.20&0.15&0.14\\
            \midrule
            \multirow{3}{*}{\begin{sideways}{ \textbf{MTR}}\end{sideways}}&\bert{}&3.04&$\textbf{2.82}$&3.41&\underline{$\textbf{2.42}$}&3.30&2.93\\
            &\roberta{}&$\textbf{2.65}$&3.80&\underline{$\textbf{2.33}$}&2.60&$\textbf{2.67}$&$\textbf{2.82}$\\
            &\albert{}&3.20&3.23&\underline{2.45}&2.58&3.17&3.18\\
            \midrule
            \multirow{3}{*}{\begin{sideways}{ \textbf{NME}}\end{sideways}}&\bert{}&\underline{0.23}&0.38&0.27&0.26&0.32&0.36\\
            &\roberta{}&\underline{$\textbf{0.18}$}&$\textbf{0.28}$&$\textbf{0.22}$&$\textbf{0.21}$&$\textbf{0.24}$&$\textbf{0.27}$\\
            &\albert{}&0.40&0.50&0.41&\underline{0.39}&0.42&0.50\\
            \bottomrule 
        \end{tabular}
        }%
    }
    \caption{Metrical evaluation (MP@1:\ mean precision at 1,\\ MTR:\ mean target rank, MNE:\ mean normalized entropy)}
    \label{tab:anteprecis}
    \end{subtable}%
    ~    
    \begin{subtable}{.49\textwidth}{
        \resizebox{.95\textwidth}{!}{%
        \begin{tabular}{llrrrrrr}
            \toprule
            &&\multicolumn{3}{c}{\textbf{objRC}}&\multicolumn{3}{c}{\textbf{subjRC}}\\	
            \midrule
            &	&who	&which	&that&	who	&which&	that\\
            \midrule
            \multirow{3}{*}{\begin{sideways}\textbf{AN}\end{sideways}}&\bert{}	&0.97&	0.94&	$\textbf{0.95}$&	\underline{0.98}&	0.94&	\underline{0.98}\\
            &\roberta{}&	\underline{$\textbf{1.00}$}&	0.97&	$\textbf{0.95}$&	\underline{$\textbf{1.00}$}&	\underline{$\textbf{1.00}$}&	$\textbf{0.96}$ \\
            &\albert&	\underline{$\textbf{1.00}$}&\underline{$\textbf{1.00}$}	 &0.90	 &0.96	&\underline{$\textbf{1.00}$}	&$\textbf{0.96}$ \\
            \midrule
            \multirow{3}{*}{\begin{sideways}\textbf{PL}\end{sideways}}&\bert{} & $\textbf{0.97}$&	0.94&	0.95&	\underline{$\textbf{1.00}$}&	0.90&	$\textbf{0.96}$ \\
            &\roberta{}& $\textbf{0.97}$&	$\textbf{0.97}$&	\underline{$\textbf{1.00}$}&	0.98&	$\textbf{0.92}$&	0.92 \\
            &\albert{} & \underline{$\textbf{0.97}$}&\underline{$\textbf{0.97}$}	 &	0.76 &	 0.96&0.65	 &	0.73 \\
            \midrule
            \multirow{3}{*}{\begin{sideways}\textbf{GR}{}\end{sideways}}& \bert{} & \underline{$\textbf{1.00}$} &	0.97 &	\underline{$\textbf{1.00}$} &	\underline{$\textbf{1.00}$} &	0.98 &	\underline{$\textbf{1.00}$} \\
            &\roberta{} & \underline{$\textbf{1.00}$}&	\underline{$\textbf{1.00}$}&	\underline{$\textbf{1.00}$}&	\underline{$\textbf{1.00}$}&	\underline{$\textbf{1.00}$}&	\underline{$\textbf{1.00}$} \\
            &\albert{} &\underline{$\textbf{1.00}$} &\underline{$\textbf{1.00}$}	 &0.95	 &	0.96 &	0.98 &0.94	 \\
            \bottomrule 
        \end{tabular}
        }%
    }
    \caption{Semantic (AN:\ animacy, PL:\ plausibility) and \\ grammaticality (GR) evaluation.}
    \label{tab:testanym}
    \end{subtable}%
    \vspace{25pt}     
    \caption{\textit{Antecedent} prediction: quantitative (a) and qualitative (b) evaluation. \textbf{Bold}:\ best result for each metric and category across models.\ \underline{Underline}:\ best result for each model per metric across categories.}
    \label{tab:antecedent}
\end{table}

\subsubsection{Antecedent Prediction}

Here, we test all three models by masking the antecedent (see example \ref{ex:m1}), considering 
again, mean precision  (MP@1), mean target rank (MTR), and normalized mean entropy (NME) (Table~\ref{tab:anteprecis}). Due to higher variation in lexical choices for antecedents, MP@1 is expectedly much lower than for relativizers. Best @1 predictions are achieved by all models for \textit{who} and \textit{that} object RCs. Comparing models, \bert{} achieves best performance in all cases. MTR is around $2.5$ and $3.5$, with antecedents in \textit{who} subject RCs being predicted most easily by all models (rank around $2.5$), also reflected in NME.  The models are equally confident in \textit{who} object RCs, while they are  less confident for the other cases, with \albert{} being the least confident and \roberta{} the most confident model. 

\ex. \label{ex:m1}{\sl Rheumatologists have to be medical detectives, because so many of the [MASK] \textbf{that} we treat are obscure. (object RC, masked target:\ \textbf{diseases})}

\noindent Considering the first predicted antecedent, we qualitatively evaluate whether (a)  the animacy constraint is met, i.e.\ \textit{who} RCs should have animate antecedents, and \textit{that} and \textit{which} most prominently inanimate ones\footnote{In cases where an elliptical version or other  is predicted, we still check whether the animacy was kept or changed.} (see~\ref{ex:ell}), 
(b) the sentence is plausible, (c) the sentence is grammatical (see example \ref{ex:gr} being ungrammatical in noun-verb agreement of \textit{animals} (plural) and \textit{was} (singular)\footnote{Here, BERT seems to consider \textit{inertia} as the antecedent, matching the agreement with \textit{was}.}). 

\ex. \label{ex:ell} {\sl They picked a great foreman in the middle-aged, an African-American [MASK] \textbf{who} she said is a science professor with a Ph.D.  (target = man, prediction = comma, animate)}

\ex. \label{ex:gr} {\sl By the time Ganivet drafted Idearium español, he was applying Ribot's psychological theories to the apparent inertia of [MASK] \textbf{which} he deemed was suffering from a collective psychological malady.  (target = Spain, prediction = animals)}

\noindent Overall, \bert{} and \roberta{} show very high accuracy for these cases (Table~\ref{tab:testanym}). In  few cases, the animacy of the antecedent is  changed (see example \ref{ex:antchanged}) and sentences are grammatical but semantically implausible or ungrammatical and implausible (see example \ref{ex:antchanged} and example \ref{ex:falseplaus}). 
\albert{} performs worst  considering plausibility,  especially for \textit{which} in subject RCs (see example \ref{ex:albert}), most cases being repetitions of words in the sentence 
(see example \ref{ex:albert-rep}). 

\ex. \label{ex:antchanged} {\sl The only thing that brightened the gloom stretching out before him was the goodness of the [MASK] \textbf{that} had offered him refuge before his trial. (target = family, prediction = darkness)}

\ex. \label{ex:falseplaus} {\sl Coffee, the cash, and the goods they purchased, even when used in traditional exchange, were devoid of the social relations with [MASK] \textbf{which} were present in traditional products. (target = predecessors, prediction = humans)}

\ex. \label{ex:albert} {\sl He and Carolyn saw all of Kevin's college [MASK] \textbf{that} were within a day's drive of McIntyre. (target = games, prediction = campuses)}

\ex. \label{ex:albert-rep} {\sl The bill makes it illegal to adopt or enforce any law or [MASK] \textbf{which} allows gays to claim discrimination. (target = policy, prediction = law)}

\noindent In summary, all models have problems with \textit{which} relativizer,  
while they perform best with \textit{who}. For animacy and plausibility the models suffer the most, especially for subject RCs, but perform quite well for grammaticality, indicating a better awareness for grammatical than semantic knowledge. 

We further evaluate the models' semantic knowledge  considering predicted types of antecedents  ((a) identical to the target, (b) synonym, (c) hypernym, general noun, determiner, pronoun, (d) hyponym\footnote{Hyponyms are semantically lower in hierarchy and thus more specific.}, or (e) not directly related (i.e.\ no direct hierarchical relationship) or completely unrelated\footnote{We relied on WordNet whenever possible (low coverage) and on other linguistic resources plus linguistic know-how of two linguists. The annotated data will be made available upon publication.}). 
Results (see Table~\ref{tab:testsem}, §\ref{sec:ANTtype}) show that all models perform less well on subject than object RCs. \albert{}  performs worst as the majority falls into not directly related/unrelated antecedents or hypernyms (more general words). \roberta{} is quite good at predicting \textit{identical} targets outperforming the other two models, especially in object RCs with \textit{who}.
 In \textit{who} subject RCs,  \bert{} and \albert{}  most often predict more general antecedents given more specific targets (hypernyms, e.g., \textit{person} instead of \textit{girl}, \textit{workers}, etc.). 
 
\section{Discussion and Conclusion}
With our work we moved towards tackling some issues in the evaluation of the linguistic capabilities of pre-trained transformer-based masked language models as, e.g., proposed by \citet{Rogers-etal2020}.Most prominently, we try to better understand how a strong performance on supervised probing tasks is reflected in the predictions of the language models. To do so, we create a dataset based on naturalistic (not artificially generated)
data and perform an extensive evaluation of masked language predictions in the context of RCs. Moreover, rather than considering  one model only, we compare three models to investigate the extend to which findings for BERT can be generalized to other transformer-based models. 

Our probing results show a significant improvement of 
all three transformer-based models over the baselines for almost all layers, suggesting indeed encoding of linguistic knowledge relevant for grammaticality classification, as shown in  previous work (e.g.\ \citet{Goldberg2019}). Performance is similar across models, with BERT and RoBERTa performing slightly better than ALBERT, and while contextualization improves overall, we have shown it helps immensely when considering less frequent RC modification types such as \texttt{which $\rightarrow$ who}, for which uncontextualized baselines learn simple heuristics. Evaluation on a diagnostic set, however, clearly reveals weaknesses of probing classifiers and model-specific behavior. \roberta{} is quite confident in wrong agreement cases between antecedent and relativizer, while this is problematic for \bert{} and \albert{}. Considering distance between relativizer and antecedent/RC verb, they clearly outperform \roberta{}. Based on these insights, \textbf{we conclude that viewing probing results in isolation can  lead to overestimating the linguistic capabilities of a model}.

Our masked language modeling evaluation provided deeper insights into model-specific differences. We evaluated relativizer as well as antecedent prediction. Overall, all models show better performance on grammatical than semantic knowledge (animacy and plausibility). Regarding relativizer prediction, all models perform worst on the target word \textit{which} (plausible, as it is the most versatile of the relativizers).
Comparing models, \bert{} is best in predicting the actual targets,  \roberta{} outperforms the others in capturing  grammatical and semantic knowledge, while \albert{} performs worst overall.

Evaluation on semantic types of antecedents shows prediction of unrelated or not directly related antecedents, especially for \albert{} in \textit{which} RCs. Interestingly, both \bert{} and \albert{} predict hierarchically more general antecedents in \textit{who} RCs, while \roberta{} is  able to better capture specificity.

We believe that more work in this direction will lead to (a) awareness of the complexity of linguistic knowledge that such models might have to capture, considering e.g.\ generalization tasks, and (b) provide enhancement in evaluation strategies on how to best capture linguistic knowledge, such as a combination of probing, diagnostic, and cloze tests, but also in developing best practices of evaluation.

\section*{Acknowledgements}

We thank the anonymous reviewers for their
valuable comments. This work was funded by the Deutsche Forschungsgemeinschaft (DFG, German Research Foundation) – Project-ID 232722074 – SFB 1102. 

\bibliography{bib}

\begin{thebibliography}{36}
\providecommand{\natexlab}[1]{#1}
\providecommand{\url}[1]{\texttt{#1}}
\expandafter\ifx\csname urlstyle\endcsname\relax
  \providecommand{\doi}[1]{doi: #1}\else
  \providecommand{\doi}{doi: \begingroup \urlstyle{rm}\Url}\fi

\bibitem[Adi et~al.(2016)Adi, Kermany, Belinkov, Lavi, and
  Goldberg]{adi2016fine}
Yossi Adi, Einat Kermany, Yonatan Belinkov, Ofer Lavi, and Yoav Goldberg.
\newblock Fine-grained analysis of sentence embeddings using auxiliary
  prediction tasks.
\newblock \emph{arXiv preprint arXiv:1608.04207}, 2016.

\bibitem[Akbik et~al.(2018)Akbik, Blythe, and Vollgraf]{akbik2018coling}
Alan Akbik, Duncan Blythe, and Roland Vollgraf.
\newblock Contextual string embeddings for sequence labeling.
\newblock In \emph{{COLING} 2018, 27th International Conference on
  Computational Linguistics}, pages 1638--1649, 2018.

\bibitem[Biber et~al.(1999)Biber, Johansson, Leech, Conrad, and
  Finegan]{Biber-etal1999}
Douglas Biber, Stig Johansson, Geoffrey Leech, Susan Conrad, and Edward
  Finegan.
\newblock \emph{{Longman Grammar of Spoken and Written English}}.
\newblock Longman, Harlow, UK, 1999.

\bibitem[Bojanowski et~al.(2016)Bojanowski, Grave, Joulin, and
  Mikolov]{bojanowski2016enriching}
Piotr Bojanowski, Edouard Grave, Armand Joulin, and Tomas Mikolov.
\newblock {Enriching Word Vectors with Subword Information}.
\newblock \emph{arXiv preprint arXiv:1607.04606}, 2016.

\bibitem[Conneau et~al.(2018)Conneau, Kruszewski, Lample, Barrault, and
  Baroni]{conneau-etal-2018-cram}
Alexis Conneau, German Kruszewski, Guillaume Lample, Lo{\"\i}c Barrault, and
  Marco Baroni.
\newblock What you can cram into a single {\$}{\&}!{\#}* vector: Probing
  sentence embeddings for linguistic properties.
\newblock In \emph{Proceedings of the 56th Annual Meeting of the Association
  for Computational Linguistics (Volume 1: Long Papers)}, pages 2126--2136,
  Melbourne, Australia, July 2018. Association for Computational Linguistics.
\newblock \doi{10.18653/v1/P18-1198}.
\newblock URL \url{https://www.aclweb.org/anthology/P18-1198}.

\bibitem[D'Arcy and Tagliamonte(2010)]{DarcyTagliamonte2010}
Alexandra D'Arcy and Sali~A. Tagliamonte.
\newblock Prestige, accommodation, and the legacy of relative who.
\newblock \emph{Language in Society}, 39\penalty0 (3):\penalty0 383–410,
  2010.
\newblock \doi{10.1017/S0047404510000205}.

\bibitem[Davies(2015)]{davies}
Mark Davies.
\newblock {Corpus of Contemporary American English (COCA)}, 2015.
\newblock URL \url{https://doi.org/10.7910/DVN/AMUDUW}.

\bibitem[Devlin et~al.(2019)Devlin, Chang, Lee, and
  Toutanova]{devlin-etal-2019-bert}
Jacob Devlin, Ming-Wei Chang, Kenton Lee, and Kristina Toutanova.
\newblock {BERT}: Pre-training of deep bidirectional transformers for language
  understanding.
\newblock In \emph{Proceedings of the 2019 Conference of the North {A}merican
  Chapter of the Association for Computational Linguistics: Human Language
  Technologies, Volume 1 (Long and Short Papers)}, pages 4171--4186,
  Minneapolis, Minnesota, June 2019. Association for Computational Linguistics.
\newblock \doi{10.18653/v1/N19-1423}.
\newblock URL \url{https://www.aclweb.org/anthology/N19-1423}.

\bibitem[Ettinger(2020)]{ettinger-2020}
Allyson Ettinger.
\newblock What bert is not: Lessons from a new suite of psycholinguistic
  diagnostics for language models.
\newblock \emph{Transactions of the Association for Computational Linguistics},
  8:\penalty0 34--48, 2020.
\newblock \doi{10.1162/tacl_a_00298}.
\newblock URL \url{https://doi.org/10.1162/tacl_a_00298}.

\bibitem[Goldberg(2019)]{Goldberg2019}
Yoav Goldberg.
\newblock {Assessing BERT's syntactic abilities}.
\newblock \emph{arXiv preprint arXiv:1901.05287}, 2019.

\bibitem[Hu et~al.(2020{\natexlab{a}})Hu, Chen, and Levy]{Hu2020ACL}
Jennifer Hu, Sherry~Yong Chen, and Roger Levy.
\newblock A closer look at the performance of neural language models on
  reflexive anaphor licensing.
\newblock In \emph{Proceedings of the Society for Computation in Linguistics
  2020}, pages 323--333, New York, New York, January 2020{\natexlab{a}}.
  Association for Computational Linguistics.
\newblock URL \url{https://www.aclweb.org/anthology/2020.scil-1.39}.

\bibitem[Hu et~al.(2020{\natexlab{b}})Hu, Gauthier, Qian, Wilcox, and
  Levy]{hu-etal-2020-systematic}
Jennifer Hu, Jon Gauthier, Peng Qian, Ethan Wilcox, and Roger Levy.
\newblock A systematic assessment of syntactic generalization in neural
  language models.
\newblock In \emph{Proceedings of the 58th Annual Meeting of the Association
  for Computational Linguistics}, pages 1725--1744, Online, July
  2020{\natexlab{b}}. Association for Computational Linguistics.
\newblock URL \url{https://www.aclweb.org/anthology/2020.acl-main.158}.

\bibitem[Huddleston and Pullum(2002)]{HuddlestonPullum2002}
Rodney Huddleston and Geoffrey~K. Pullum.
\newblock \emph{The Cambridge Grammar of the English Language}.
\newblock Cambridge University Press, 2002.
\newblock \doi{10.1017/9781316423530}.

\bibitem[Jiang et~al.(2020)Jiang, Xu, Araki, and Neubig]{jiang2019can}
Zhengbao Jiang, Frank~F. Xu, Jun Araki, and Graham Neubig.
\newblock How can we know what language models know?
\newblock \emph{Transactions of the Association for Computational Linguistics},
  8:\penalty0 423--438, 2020.
\newblock \doi{10.1162/tacl_a_00324}.
\newblock URL \url{https://www.aclweb.org/anthology/2020.tacl-1.28}.

\bibitem[Kassner and Sch{\"u}tze(2020)]{kassner-schutze-2020-negated}
Nora Kassner and Hinrich Sch{\"u}tze.
\newblock Negated and misprimed probes for pretrained language models: Birds
  can talk, but cannot fly.
\newblock In \emph{Proceedings of the 58th Annual Meeting of the Association
  for Computational Linguistics}, pages 7811--7818, Online, July 2020.
  Association for Computational Linguistics.
\newblock URL \url{https://www.aclweb.org/anthology/2020.acl-main.698}.

\bibitem[Kim et~al.(2019)Kim, Patel, Poliak, Xia, Wang, McCoy, Tenney, Ross,
  Linzen, Van~Durme, Bowman, and Pavlick]{Kim-etal2019}
Najoung Kim, Roma Patel, Adam Poliak, Patrick Xia, Alex Wang, Tom McCoy, Ian
  Tenney, Alexis Ross, Tal Linzen, Benjamin Van~Durme, Samuel~R. Bowman, and
  Ellie Pavlick.
\newblock Probing what different {NLP} tasks teach machines about function word
  comprehension.
\newblock In \emph{Proceedings of the Eighth Joint Conference on Lexical and
  Computational Semantics (*{SEM} 2019)}, pages 235--249, Minneapolis,
  Minnesota, June 2019. Association for Computational Linguistics.
\newblock \doi{10.18653/v1/S19-1026}.
\newblock URL \url{https://www.aclweb.org/anthology/S19-1026}.

\bibitem[Lan et~al.(2020)Lan, Chen, Goodman, Gimpel, Sharma, and
  Soricut]{Lan2020ALBERT:}
Zhenzhong Lan, Mingda Chen, Sebastian Goodman, Kevin Gimpel, Piyush Sharma, and
  Radu Soricut.
\newblock Albert: A lite bert for self-supervised learning of language
  representations.
\newblock In \emph{International Conference on Learning Representations}, 2020.
\newblock URL \url{https://openreview.net/forum?id=H1eA7AEtvS}.

\bibitem[Liu et~al.(2019{\natexlab{a}})Liu, Gardner, Belinkov, Peters, and
  Smith]{liu-etal-2019-linguistic}
Nelson~F. Liu, Matt Gardner, Yonatan Belinkov, Matthew~E. Peters, and Noah~A.
  Smith.
\newblock Linguistic knowledge and transferability of contextual
  representations.
\newblock In \emph{Proceedings of the 2019 Conference of the North {A}merican
  Chapter of the Association for Computational Linguistics: Human Language
  Technologies, Volume 1 (Long and Short Papers)}, pages 1073--1094,
  Minneapolis, Minnesota, June 2019{\natexlab{a}}. Association for
  Computational Linguistics.
\newblock \doi{10.18653/v1/N19-1112}.
\newblock URL \url{https://www.aclweb.org/anthology/N19-1112}.

\bibitem[Liu et~al.(2019{\natexlab{b}})Liu, Ott, Goyal, Du, Joshi, Chen, Levy,
  Lewis, Zettlemoyer, and Stoyanov]{liu2019roberta}
Yinhan Liu, Myle Ott, Naman Goyal, Jingfei Du, Mandar Joshi, Danqi Chen, Omer
  Levy, Mike Lewis, Luke Zettlemoyer, and Veselin Stoyanov.
\newblock Roberta: A robustly optimized bert pretraining approach.
\newblock \emph{arXiv preprint arXiv:1907.11692}, 2019{\natexlab{b}}.

\bibitem[Marvin and Linzen(2018)]{marvin-linzen-2018-targeted}
Rebecca Marvin and Tal Linzen.
\newblock Targeted syntactic evaluation of language models.
\newblock In \emph{Proceedings of the 2018 Conference on Empirical Methods in
  Natural Language Processing}, pages 1192--1202, Brussels, Belgium,
  October-November 2018. Association for Computational Linguistics.
\newblock \doi{10.18653/v1/D18-1151}.
\newblock URL \url{https://www.aclweb.org/anthology/D18-1151}.

\bibitem[Mikolov et~al.(2018)Mikolov, Grave, Bojanowski, Puhrsch, and
  Joulin]{mikolov2018advances}
Tomas Mikolov, Edouard Grave, Piotr Bojanowski, Christian Puhrsch, and Armand
  Joulin.
\newblock Advances in pre-training distributed word representations.
\newblock In \emph{Proceedings of the International Conference on Language
  Resources and Evaluation (LREC 2018)}, 2018.

\bibitem[Pedregosa et~al.(2011)Pedregosa, Varoquaux, Gramfort, Michel, Thirion,
  Grisel, Blondel, Prettenhofer, Weiss, Dubourg, Vanderplas, Passos,
  Cournapeau, Brucher, Perrot, and Duchesnay]{scikit-learn}
F.~Pedregosa, G.~Varoquaux, A.~Gramfort, V.~Michel, B.~Thirion, O.~Grisel,
  M.~Blondel, P.~Prettenhofer, R.~Weiss, V.~Dubourg, J.~Vanderplas, A.~Passos,
  D.~Cournapeau, M.~Brucher, M.~Perrot, and E.~Duchesnay.
\newblock Scikit-learn: {M}achine {L}earning in {P}ython.
\newblock \emph{Journal of Machine Learning Research}, 12:\penalty0 2825--2830,
  2011.

\bibitem[Pennington et~al.(2014)Pennington, Socher, and
  Manning]{pennington-etal-2014-glove}
Jeffrey Pennington, Richard Socher, and Christopher Manning.
\newblock {G}lo{V}e: Global vectors for word representation.
\newblock In \emph{Proceedings of the 2014 Conference on Empirical Methods in
  Natural Language Processing ({EMNLP})}, pages 1532--1543, Doha, Qatar,
  October 2014. Association for Computational Linguistics.
\newblock \doi{10.3115/v1/D14-1162}.
\newblock URL \url{https://www.aclweb.org/anthology/D14-1162}.

\bibitem[Petroni et~al.(2019)Petroni, Rockt{\"a}schel, Riedel, Lewis, Bakhtin,
  Wu, and Miller]{petroni-etal-2019-language}
Fabio Petroni, Tim Rockt{\"a}schel, Sebastian Riedel, Patrick Lewis, Anton
  Bakhtin, Yuxiang Wu, and Alexander Miller.
\newblock Language models as knowledge bases?
\newblock In \emph{Proceedings of the 2019 Conference on Empirical Methods in
  Natural Language Processing and the 9th International Joint Conference on
  Natural Language Processing (EMNLP-IJCNLP)}, pages 2463--2473, Hong Kong,
  China, November 2019. Association for Computational Linguistics.
\newblock \doi{10.18653/v1/D19-1250}.
\newblock URL \url{https://www.aclweb.org/anthology/D19-1250}.

\bibitem[Pruksachatkun et~al.(2020)Pruksachatkun, Phang, Liu, Htut, Zhang,
  Pang, Vania, Kann, and Bowman]{pruksachatkun-etal-2020-intermediate}
Yada Pruksachatkun, Jason Phang, Haokun Liu, Phu~Mon Htut, Xiaoyi Zhang,
  Richard~Yuanzhe Pang, Clara Vania, Katharina Kann, and Samuel~R. Bowman.
\newblock Intermediate-task transfer learning with pretrained language models:
  When and why does it work?
\newblock In \emph{Proceedings of the 58th Annual Meeting of the Association
  for Computational Linguistics}, pages 5231--5247, Online, July 2020.
  Association for Computational Linguistics.
\newblock URL \url{https://www.aclweb.org/anthology/2020.acl-main.467}.

\bibitem[Quirk(1957)]{Quirk1957}
Randolph Quirk.
\newblock {Relative Clauses in Educated Spoken English}.
\newblock \emph{English Studies}, 38:\penalty0 97--109, 1957.

\bibitem[Rogers et~al.(2020)Rogers, Kovaleva, and Rumshisky]{Rogers-etal2020}
Anna Rogers, Olga Kovaleva, and Anna Rumshisky.
\newblock A primer in bertology: What we know about how bert works, 2020.

\bibitem[Talmor et~al.(2020)Talmor, Tafjord, Clark, Goldberg, and
  Berant]{talmor2020teaching}
Alon Talmor, Oyvind Tafjord, Peter Clark, Yoav Goldberg, and Jonathan Berant.
\newblock {Teaching Pre-Trained Models to Systematically Reason Over Implicit
  Knowledge}.
\newblock \emph{arXiv preprint arXiv:2006.06609}, 2020.

\bibitem[Tenney et~al.(2019)Tenney, Xia, Chen, Wang, Poliak, McCoy, Kim, Durme,
  Bowman, Das, and Pavlick]{tenney2018what}
Ian Tenney, Patrick Xia, Berlin Chen, Alex Wang, Adam Poliak, R~Thomas McCoy,
  Najoung Kim, Benjamin~Van Durme, Sam Bowman, Dipanjan Das, and Ellie Pavlick.
\newblock What do you learn from context? probing for sentence structure in
  contextualized word representations.
\newblock In \emph{International Conference on Learning Representations}, 2019.
\newblock URL \url{https://openreview.net/forum?id=SJzSgnRcKX}.

\bibitem[Vaswani et~al.(2017)Vaswani, Shazeer, Parmar, Uszkoreit, Jones, Gomez,
  Kaiser, and Polosukhin]{NIPS2017_7181}
Ashish Vaswani, Noam Shazeer, Niki Parmar, Jakob Uszkoreit, Llion Jones,
  Aidan~N Gomez, \L~ukasz Kaiser, and Illia Polosukhin.
\newblock Attention is all you need.
\newblock In I.~Guyon, U.~V. Luxburg, S.~Bengio, H.~Wallach, R.~Fergus,
  S.~Vishwanathan, and R.~Garnett, editors, \emph{Advances in Neural
  Information Processing Systems 30}, pages 5998--6008. Curran Associates,
  Inc., 2017.
\newblock URL
  \url{http://papers.nips.cc/paper/7181-attention-is-all-you-need.pdf}.

\bibitem[Wang et~al.(2018)Wang, Singh, Michael, Hill, Levy, and
  Bowman]{wang-etal-2018-glue}
Alex Wang, Amanpreet Singh, Julian Michael, Felix Hill, Omer Levy, and Samuel
  Bowman.
\newblock {GLUE}: A multi-task benchmark and analysis platform for natural
  language understanding.
\newblock In \emph{Proceedings of the 2018 {EMNLP} Workshop {B}lackbox{NLP}:
  Analyzing and Interpreting Neural Networks for {NLP}}, pages 353--355,
  Brussels, Belgium, November 2018. Association for Computational Linguistics.
\newblock \doi{10.18653/v1/W18-5446}.
\newblock URL \url{https://www.aclweb.org/anthology/W18-5446}.

\bibitem[Warstadt and Bowman(2019)]{WarstadtBowman2019}
Alex Warstadt and Samuel~R. Bowman.
\newblock Linguistic analysis of pretrained sentence encoders with
  acceptability judgments, 2019.

\bibitem[Warstadt et~al.(2019)Warstadt, Cao, Grosu, Peng, Blix, Nie, Alsop,
  Bordia, Liu, Parrish, Wang, Phang, Mohananey, Htut, Jeretic, and
  Bowman]{warstadt-etal-2019-investigating}
Alex Warstadt, Yu~Cao, Ioana Grosu, Wei Peng, Hagen Blix, Yining Nie, Anna
  Alsop, Shikha Bordia, Haokun Liu, Alicia Parrish, Sheng-Fu Wang, Jason Phang,
  Anhad Mohananey, Phu~Mon Htut, Paloma Jeretic, and Samuel~R. Bowman.
\newblock Investigating {BERT}{'}s knowledge of language: Five analysis methods
  with {NPI}s.
\newblock In \emph{Proceedings of the 2019 Conference on Empirical Methods in
  Natural Language Processing and the 9th International Joint Conference on
  Natural Language Processing (EMNLP-IJCNLP)}, pages 2877--2887, Hong Kong,
  China, November 2019. Association for Computational Linguistics.
\newblock \doi{10.18653/v1/D19-1286}.
\newblock URL \url{https://www.aclweb.org/anthology/D19-1286}.

\bibitem[Warstadt et~al.(2020)Warstadt, Parrish, Liu, Mohananey, Peng, Wang,
  and Bowman]{warstadt2019blimp}
Alex Warstadt, Alicia Parrish, Haokun Liu, Anhad Mohananey, Wei Peng, Sheng-Fu
  Wang, and Samuel~R. Bowman.
\newblock {BL}i{MP}: The benchmark of linguistic minimal pairs for {E}nglish.
\newblock \emph{Transactions of the Association for Computational Linguistics},
  8:\penalty0 377--392, 2020.
\newblock \doi{10.1162/tacl_a_00321}.
\newblock URL \url{https://www.aclweb.org/anthology/2020.tacl-1.25}.

\bibitem[Wilcox et~al.(2019)Wilcox, Qian, Futrell, Ballesteros, and
  Levy]{wilcox-etal-2019-structural}
Ethan Wilcox, Peng Qian, Richard Futrell, Miguel Ballesteros, and Roger Levy.
\newblock Structural supervision improves learning of non-local grammatical
  dependencies.
\newblock In \emph{Proceedings of the 2019 Conference of the North {A}merican
  Chapter of the Association for Computational Linguistics: Human Language
  Technologies, Volume 1 (Long and Short Papers)}, pages 3302--3312,
  Minneapolis, Minnesota, June 2019. Association for Computational Linguistics.
\newblock \doi{10.18653/v1/N19-1334}.
\newblock URL \url{https://www.aclweb.org/anthology/N19-1334}.

\bibitem[Wolf et~al.(2019)Wolf, Debut, Sanh, Chaumond, Delangue, Moi, Cistac,
  Rault, Louf, Funtowicz, et~al.]{wolf2019huggingface}
Thomas Wolf, Lysandre Debut, Victor Sanh, Julien Chaumond, Clement Delangue,
  Anthony Moi, Pierric Cistac, Tim Rault, R{\'e}mi Louf, Morgan Funtowicz,
  et~al.
\newblock {HuggingFace's Transformers: State-of-the-art Natural Language
  Processing}.
\newblock \emph{ArXiv}, pages arXiv--1910, 2019.

\end{thebibliography}
\bibliographystyle{plainnat}

\newpage
\appendix
\section{Appendix}

\subsection{Probing Dataset}
\label{sec:prob_data}

\begin{figure}[h]
    \centering
    \begin{subfigure}[t]{0.53\textwidth}
        \centering
        \includegraphics[width=1.0\textwidth]{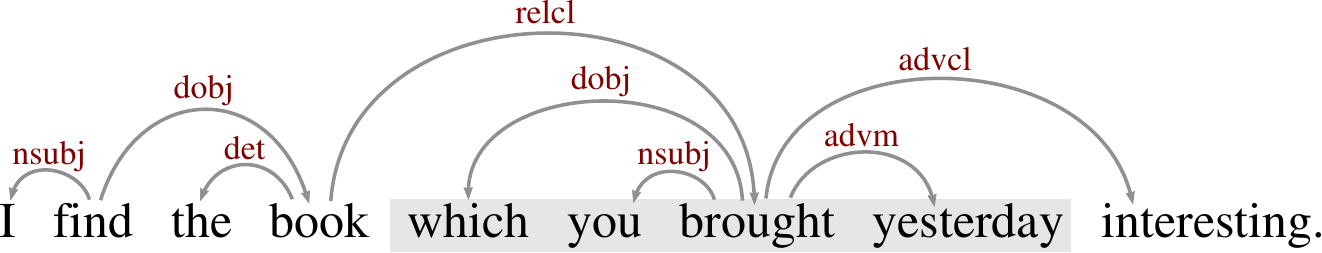}
        \caption{}
    \end{subfigure}%
    ~ 
    \begin{subfigure}[t]{0.47\textwidth}
        \centering
        \includegraphics[width=1.0\textwidth]{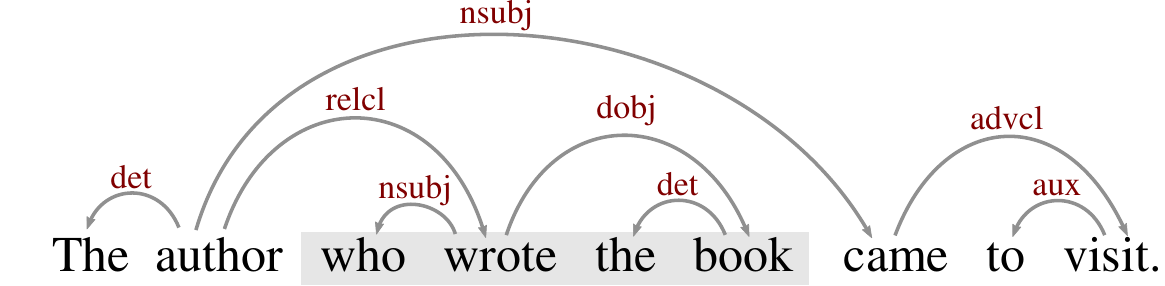}
        \caption{}
    \end{subfigure}%
    \vspace{15pt}
    \caption{Visualized dependency parse trees extracted by SpaCy: (a) object RC, and (b) subject RC.}
    \vspace{25pt}
    \label{fig:appendix_data_creation}
\end{figure}

Fig.~\ref{fig:appendix_data_creation} shows how to identify sentences containing RCs using a dependency parse tree, which were obtained using SpaCy. A sentence with an RC can be identified if the antecedent has an outgoing edge with the tag \textsc{relcl}. The RC in the sentence can be further extracted since the main verb in the RC would have an incoming edge with the tag \textsc{relcl}, and the subtree where RC main verb is the head constitutes the RC. This procedure enables us to extract sentences with RCs that are grammatical from text.

\begin{figure}[h]
    \centering
    \includegraphics[width=0.85\textwidth]{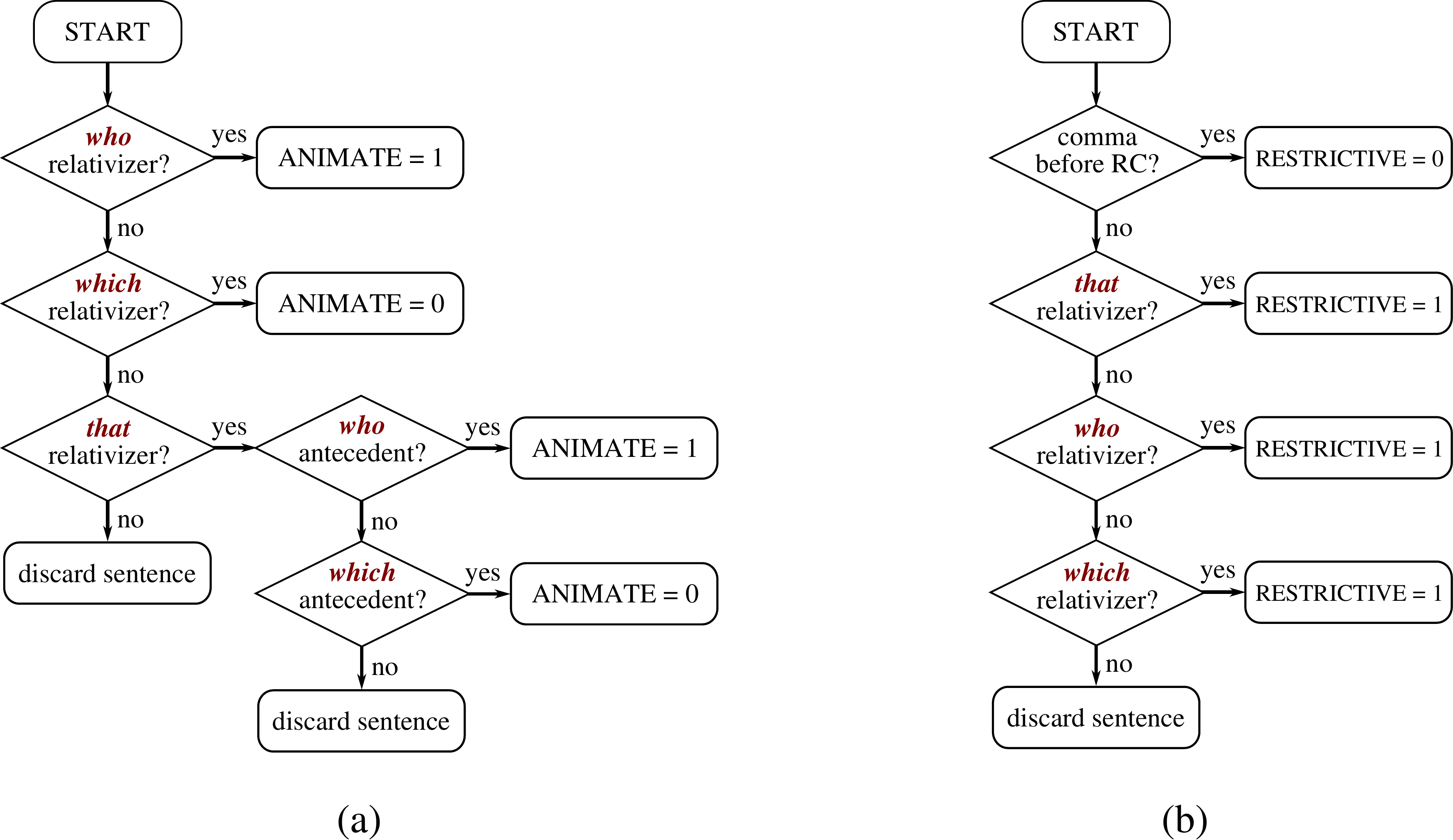}
    \caption{Annotation decision process for meta-data variables: (a) \textit{Animate}, relativizer \textbf{who} and \textbf{which} are directly categorized as animate and non-animate, respectively.  The relativizer \textbf{that} can be either way; thus, we categorize these sentences based on the antecedent. We compile two disjoints sets for antecedents that exclusively occur either with \textbf{who} or  \textbf{which}, and the decision is made based on the membership of the antecedent. If the antecedent is not a member of either set, the sentence is discarded. (b) \textit{Restrictive} can be easily identified since non-restrictive RCs in American English are always preceded by comma ``,''. }
    \label{fig:decision-trees}
    \vspace{25pt}
\end{figure}

However, probing classifiers for grammatical acceptability require both grammatical and ungrammatical sentences. Since grammaticality of RCs in English depends on  restrictiveness, animacy of the head noun, and whether the relativizer occupies the subject or the object position in the sentence, we populate three meta-data variables for each grammatical sentence with respect to these aspects of RCs: \textsc{animate}, \textsc{restrictive}, and \textsc{subjrc}.  The variables \textsc{animate} and \textsc{restrictive} are populated using a set of hand-crafted rules for the usage of RCs in American English (illustrated in Fig.\ \ref{fig:decision-trees}), while the variable \textsc{subjrc} is populated using the incoming edge to the relativizer in the parse tree (e.g.\ \textsc{nsubj} for subject RC vs.\ \textsc{dobj} for object RC). We discard all sentences where at least one meta-data variable cannot be populated with certainty.

\noindent Using the aforementioned annotation procedure, and given the values of the three meta-data variables, each sentence can be manipulated to create an ungrammatical sentence that forms a minimal pair with the original sentence. The resulting paradigms from the three meta-data variables and the set of all possible modifications that be applied to each paradigm are presented in Table \ref{tab:modifications}. Each grammatical sentence in the dataset is manipulated using all possible modifications of the paradigm to generate a ``bag-of-sentences'' associated with each sentence (where the original grammatical sentence is a member of).  To create the final dataset, we sample one sentence from each bag and construct a balanced dataset of 48,060 sentences with grammatical acceptability labels. Tables \ref{tab:stats_splits} and \ref{tab:stats_mods} show summary statistics of the dataset.

\begin{table}[t]
\resizebox{.99\textwidth}{!}{
\begin{tabular}{@{}cccccl@{}}
\toprule
\textbf{Animate}   & \textbf{Restrictive} & \textbf{Subject RC} & \textbf{Modification} & \textbf{Label} & \textbf{Example}      
  \\ \midrule
  \multirow{3}{*}{1} & \multirow{3}{*}{1}   & \multirow{3}{*}{1}  & None                  & 1              & Katrina Haus was a woman \textbf{who} sought to attract stares, not turn them away.    \\
                  &                      &                     & Relativizer omission   & 0              & *Katrina Haus was a woman sought to attract stares, not turn them away.        \\
                  &                      &                     & who $\rightarrow$ which          & 0              & *Katrina Haus was a woman \textbf{which} sought to attract stares, not turn them away. 
                  
                  \\ \midrule
  \multirow{3}{*}{1} & \multirow{3}{*}{1}   & \multirow{3}{*}{0}  & None                  & 1              & Two people \textbf{who} I loved very much have left me .    \\
                  &                      &                     & Relativizer omission   & 1              & Two people I loved very much have left me .       \\
                  &                      &                     & who $\rightarrow$ which          & 0              & *Two people \textbf{which} I loved very much have left me .
                  \\ \midrule

\multirow{3}{*}{1} & \multirow{3}{*}{0}   & \multirow{3}{*}{1}  & None                  & 1              & Buck , \textbf{who} was snoozing over in the corner , woke up and looked around.          \\
&                      &                     & Relativizer omission  & 0              & *Buck , was snoozing over in the corner , woke up and looked around.                \\
&                      &                     & who $\rightarrow$which           & 0              & *Buck , \textbf{which} was snoozing over in the corner , woke up and looked around. \\ \midrule
    
\multirow{3}{*}{1} & \multirow{3}{*}{0}   & \multirow{3}{*}{0}  & None                  & 1              & Sally turned with a welcoming grin , expecting to see Gus , \textbf{whom} she liked a lot.          \\
&                      &                     & Relativizer omission  & 0              & *Sally turned with a welcoming grin , expecting to see Gus , she liked a lot                \\
&                      &                     & who $\rightarrow$which           & 0              & *Sally turned with a welcoming grin , expecting to see Gus , \textbf{which} she liked a lot   \\ \midrule

\multirow{4}{*}{0} & \multirow{4}{*}{1}   & \multirow{4}{*}{1}  & None                  & 1              & One is a rather, um, disconcerting bit of information \textbf{which} has reached my ears.         \\
                  &                      &                     & Relativizer omission   & 0              & *One is a rather, um, disconcerting bit of information has reached my ears.              \\
                  &                      &                     & which $\rightarrow$ who          & 0              & *One is a rather, um, disconcerting bit of information \textbf{who} has reached my ears.          \\
                  &                      &                     & which $\rightarrow$ that         & 1              & One is a rather, um, disconcerting bit of information \textbf{that} has reached my ears.         

\\ \midrule

\multirow{4}{*}{0} & \multirow{4}{*}{1}   & \multirow{4}{*}{0}  & None                  & 1              & Pulls out a course catalog, various forms, and a letter \textbf{which} she hands to Kevin.          \\
                  &                      &                     & Relativizer omission   & 0              & *Pulls out a course catalog, various forms, and a letter  she hands to Kevin.             \\
                  &                      &                     & which $\rightarrow$ who          & 0              & *Pulls out a course catalog, various forms, and a letter \textbf{who} she hands to Kevin.                 \\
                  &                      &                     & which $\rightarrow$ that          & 1              & Pulls out a course catalog, various forms, and a letter \textbf{that} she hands to Kevin.    
                  
\\ \midrule
\multirow{3}{*}{0} & \multirow{3}{*}{0}   & \multirow{3}{*}{1} & None                  & 1              & I never saw a penny in royalties, \textbf{which} was all right with me.                             \\
                  &                      &                     & Relativizer omission   & 0              & *I never saw a penny in royalties, was all right with me.                                  \\
                  &                      &                     & which $\rightarrow$ who          & 0              & *I never saw a penny in royalties, \textbf{who} was all right with me.                                    
                  
\\ \midrule
\multirow{3}{*}{0} & \multirow{3}{*}{0}   & \multirow{3}{*}{0} & None                  & 1              & Lyric clips her Walkman to her fanny pack, \textbf{which} she wears pouch forward.                            \\
                  &                      &                     & Relativizer omission   & 0              & *Lyric clips her Walkman to her fanny pack, she wears pouch forward.                                 \\
                  &                      &                     & which $\rightarrow$ who          & 0              & *Lyric clips her Walkman to her fanny pack, \textbf{who} she wears pouch forward.
                  
                  \\ \bottomrule
\end{tabular}
}

\caption{Examples of the generated paradigms and the possible modifications for each paradigm. It is worth pointing out that not all sentence modifications yield an unacceptable sentence. For example, the modification \textit{relativizer omission} keeps the sentence grammatical in restrictive object RCs.  }
\label{tab:modifications}
\end{table}

\vspace{15pt}

\begin{table}[h]
\centering
\begin{tabular}{@{}rclcclcclcc@{}}
\toprule
\multirow{2}{*}{} & \multirow{2}{*}{\textbf{Total}} &  & \multicolumn{2}{c}{\textbf{Acceptable}} &  & \multicolumn{2}{c}{\textbf{Animate}} &  & \multicolumn{2}{c}{\textbf{Restrictive}} \\ \cmidrule(lr){4-5} \cmidrule(lr){7-8} \cmidrule(l){10-11} 
                  &                        &  & 0              & 1             &  & 0            & 1            &  & 0              & 1              \\ \midrule
Training          & 42720                  &  & 21360          & 21360         &  & 11106        & 31614        &  & 25947          & 16773             \\
Test              & 5340                   &  & 2670           & 2670          &  & 1439         & 3901         &  & 3229           & 2111           \\ \bottomrule
\end{tabular}
\caption{Summary statistics of the dataset splits used in this paper. The dataset is balanced with respect to acceptability judgment (the probing label \textbf{Acceptable}). \textbf{Animate} and \textbf{Restrictive} are meta-data variables, they are only used for creating the data and the modifications, not for probing. }
\label{tab:stats_splits}
\end{table}

\vspace{20pt}

\begin{table}[h]
\resizebox{1.0\textwidth}{!}{
\centering
\begin{tabular}{@{}rcccccccccccccc@{}}
\toprule
             & \multicolumn{2}{c}{No modif.} &  & \multicolumn{2}{c}{Rel. omission} &  & \multicolumn{2}{c}{who $\rightarrow$ which} &  & \multicolumn{2}{c}{which $\rightarrow$ who} &  & \multicolumn{2}{c}{which $\rightarrow$ that} \\ \cmidrule(lr){2-3} \cmidrule(lr){5-6} \cmidrule(lr){8-9} \cmidrule(lr){11-12} \cmidrule(l){14-15} 
Acceptability & 0              & 1                  &  & 0                    & 1                 &  & 0                & 1            &  & 0                & 1             &  & 0             & 1                \\ \midrule
Training     & --              & 20170              &  & 10512                & 116               &  & 8291             & --            &  & 2557             & --             &  & --             & 1074             \\
Test         & --              & 2670               &  & 1299                 & 19                &  & 1025             & --            &  & 346              & --             &  & --             & 125              \\ \bottomrule
\end{tabular}
}
\caption{Summary statistics of the dataset with respect to the modifications. Note that modifications are not balanced.}
\label{tab:stats_mods}
\end{table}

\vspace{20pt}

\subsection{Probing Results}\label{appprob}

\begin{figure}[h]
    \centering
    \begin{subfigure}[t]{.49\textwidth}
        \centering
        \includegraphics[width=1.0\textwidth]{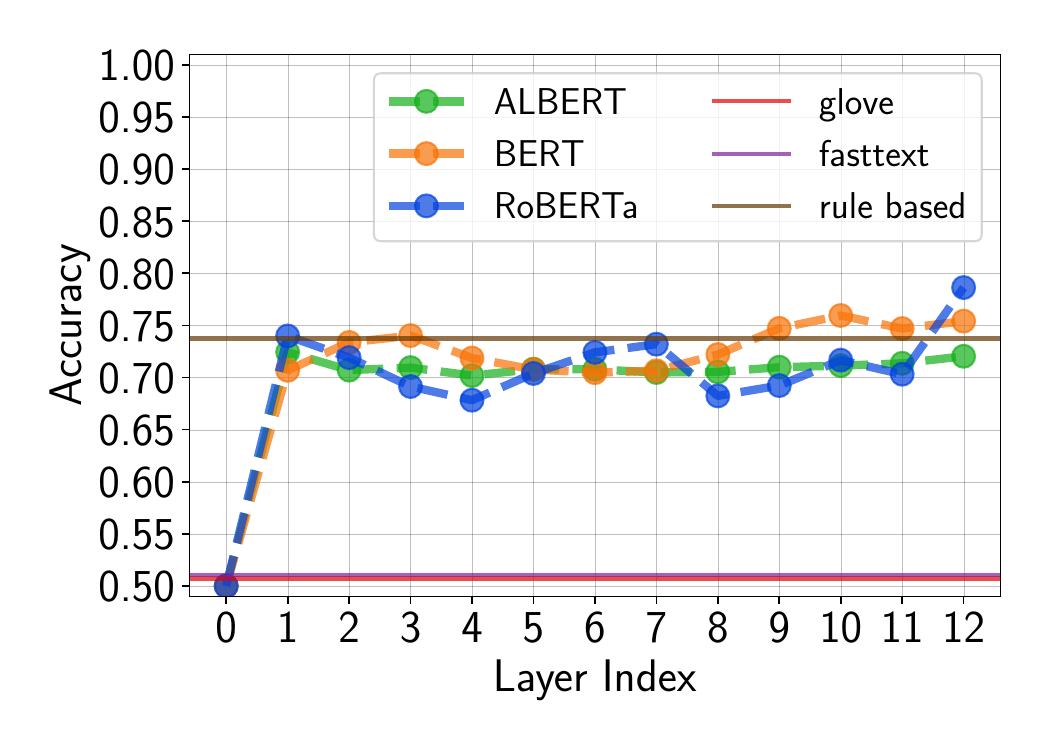}
        \caption{CLS-pooling}
        \label{fig:mean}
    \end{subfigure}%
    ~
    \begin{subfigure}[t]{.49\textwidth}
        \centering
        \includegraphics[width=1.0\textwidth]{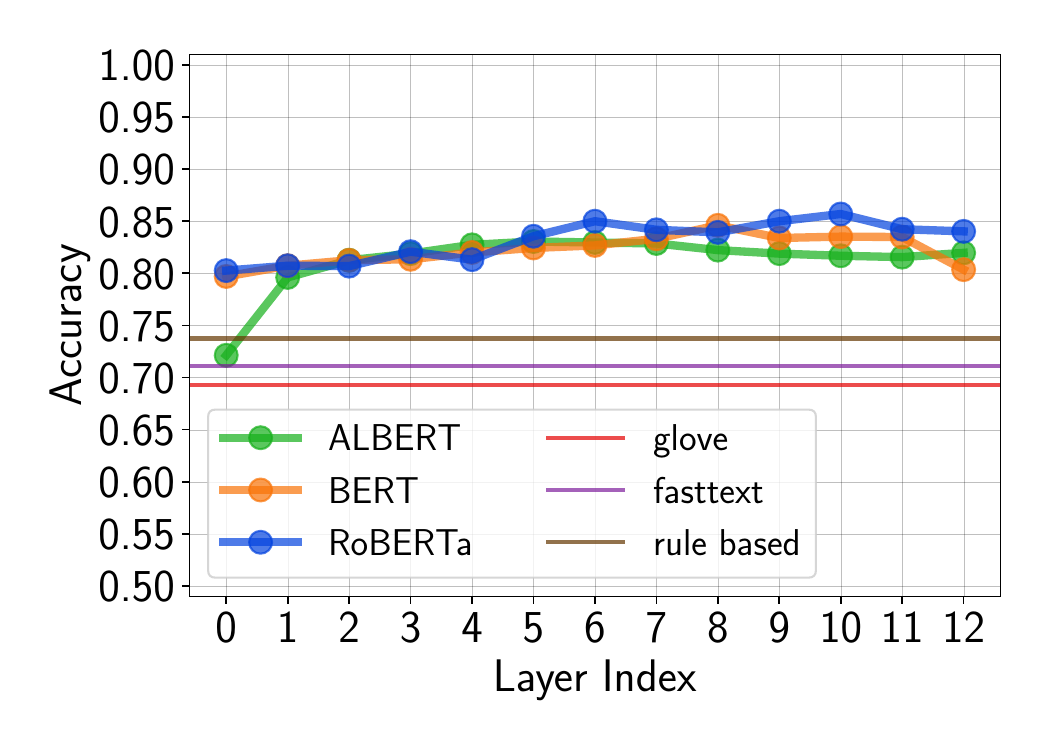}
        \caption{mean-pooling}
        \label{fig:cls}
    \end{subfigure}%
    \vspace{15pt}
    \caption{Side-by-side comparison of layer-wise probing accuracy on the test set for pre-trained transformer and baseline models using (a) CLS-pooling and (b) mean-pooling.}
    \vspace{25pt}
    \label{fig:appendix:probing_results}
\end{figure}

Fig.\ref{fig:appendix:probing_results} shows layer-wise probing accuracies for all models considered in this paper. CLS-pooling clearly leads to worse sentence representations performing on par with the non-contextualized GloVe and fasttext mean-embeddings for most layers.

\subsubsection{ALBERT-base-v1 vs. ALBERT-xxlarge-v1}
\label{sec:alberts}

\begin{figure}[h]
    \centering
        \centering
        \includegraphics[width=.49\textwidth]{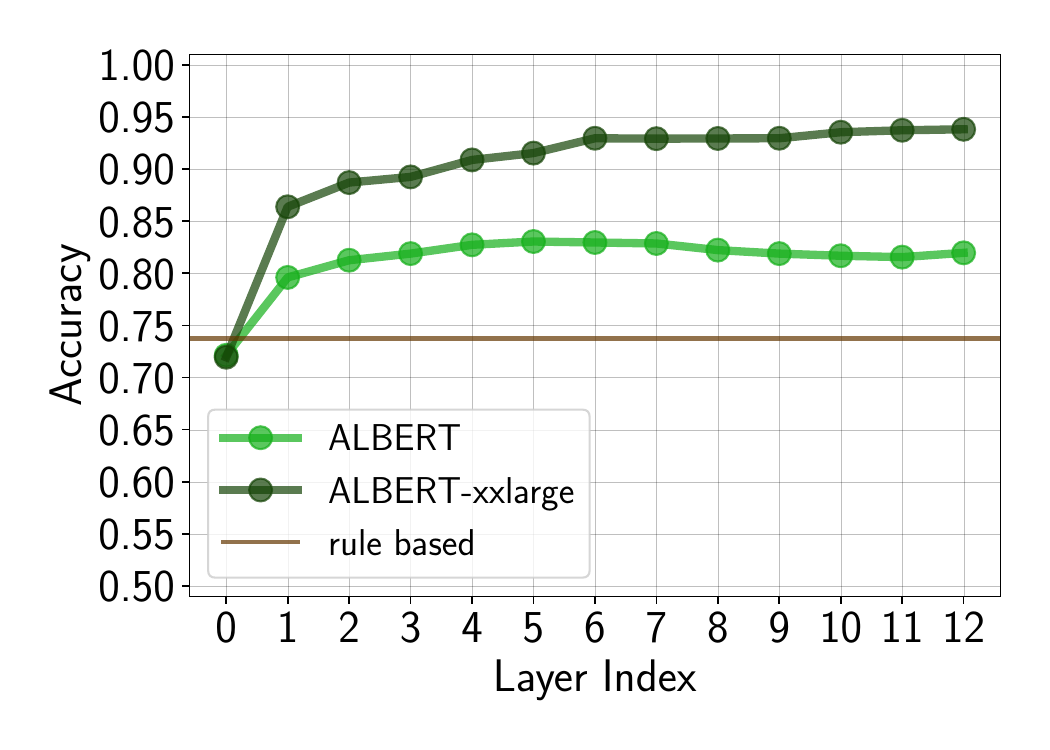}
        \label{fig:mean-albert}
    \caption{Comparison of ALBERT-base-v1 and ALBERT-xxlarge-v1 using mean-pooling. Table~\ref{tab:probing-results_appendix} shows the results of the best ALBERT-xxlarge-v1 probing classifier compared to all other models grouped by modification.}
    \vspace{25pt}
    \label{fig:appendix:probing_results_alberts}
\end{figure}

\noindent Fig.\ref{fig:appendix:probing_results_alberts} shows layer-wise probing accuracies for ALBERT-base-v1 (ALBERT) and ALBERT-xxlarge-v1. ALBERT-xxlarge-v1 contains 235M parameters and hence roughly 10x as many as ALBERT-base-v1 and 2x as many as BERT and RoBERTa. As the results in Fig.\ref{fig:appendix:probing_results_alberts} show, the larger number of parameters indeed results in a significantly higher probing accuracy, outperforming all other models by a large margin. However, it should be noted that the larger number of parameters is the results of a larger dimensionality of ALBERT-xxlarge-v1's embeddings space, which is 5x larger that that of ALBERT-base-v1, BERT, and RoBERTa (4096 vs. 768). This makes the obtained probing results not directly comparable. We leave a more detailed investigation of the role of the size of the embedding space for future work.

\begin{table}[t]
    \centering \small
        \begin{tabular}{lcccccccc}
        \toprule
              \textbf{Modification} &  \textbf{GloVE} &  \textbf{fasttext} &  \textbf{rule based} & \textbf{BERT} &  \textbf{RoBERTa} &  \textbf{ALBERT} &  \textbf{ALBERT-xxlarge} \\ \midrule
              no modification & $67.3$ & $69.2$  & $100$ & $83.7$ $(78.4)$ & $85.3$ $(79.7)$ & $83.6$ $(71.5)$ & $95.0$ $(70.6)$ \\
              relativizer omission  & $77.2$ & $85.3$ & $97.6$ & $96.0$ $(98.3)$ & $95.1$ $(98.2)$ & $96.4$ $(94.8)$ & $96.2$ $(93.5)$ \\
              who $\rightarrow$ which & $88.0$ & $81.7$ & $0.00$ & $84.9$ $(88.4)$ & $87.3$ $(89.6)$ & $78.8$ $(68.3)$ & $94.6$ $(69.0)$ \\
              which $\rightarrow$  who & $18.2$ & $21.3$ & $0.00$ & $47.9$ $(1.73)$ & $50.0$ $(0.02)$ & $44.2$ $(18.7)$ & $80.6$ $(21.3)$ \\
              which $\rightarrow$  that & $12.8$ & $8.0$ & $100$ & $80.0$ $(52.0)$ & $77.6$ $(44.8)$ & $72.0$ $(22.4)$ & $73.6$ $(33.6)$ \\ \midrule
              total &  $69.2$ &  $71.0$ & $73.7$ &  $84.6$ $(79.7)$ &  $85.5$ $(80.1)$ &  $83.0$ $(72.1)$ & $93.8$ $(71.9)$ \\ 
        \bottomrule     
        \end{tabular}
    \caption{Test accuracy (in $\%$) grouped by modification type (cf. Table~\ref{tab:stats_mods} for statistics). For BERT, RoBERTa, ALBERT-base-v1 (ALBERT) and ALBERT-xxlarge-v1 we select the best model according to the probing results shown in Fig.~\ref{fig:layer-wise-accuracy}. Numbers in parenthesis show the accuracy of the non-contextualized baseline (layer 0) for each model. ALBERT-xxlarge-v1 performs especially well on the \texttt{which $\rightarrow$ who} modification.
    }
    \label{tab:probing-results_appendix}
\end{table}

\subsection{Qualitative Analysis for Predicted Type of Antecedent}\label{sec:ANTtype}
Table~\ref{tab:testsem} shows percentage for predicted types of antecedents:\ (a) identical to  target, (b) synonym, (c) hypernym, general noun, determiner, pronoun, (d) hyponym (i.e.\ semantically lower in hierarchy and thus more specific), or (e) not directly related (i.e.\ no direct hierarchical relationship) or completely unrelated. Results show \albert{} to perform worst as the majority falls into not directly related/unrelated antecedents or hypernyms (more general words). \roberta{} is quite good at predicting \textit{identical} targets outperforming the other two models, especially in object RCs with \textit{who}. While the \textit{which} case in object RCs is harder for all models, \roberta{} still makes best predictions considering identical and synonym prediction. All models perform less well on subject RCs, especially for \textit{which} and \textit{that} (40\% to $>50\%$ of unrelated/not directly related targets). Interestingly, for animate antecedents (\textit{who} relativizer), the majority for all models falls into hypernyms, i.e.\ more general variants. This is especially the case for \bert{} and \albert{}. 
\begin{table}[h]\footnotesize
    \centering
    \begin{tabular}{llrrrrrr}
    \toprule
&\textbf{type}&\multicolumn{3}{c}{\textbf{objRC}}&\multicolumn{3}{c}{\textbf{subjRC}}\\	
\midrule
&	&who	&which	&that&	who	&which&	that\\
\midrule
\multirow{5}{*}{\begin{sideways}\textbf{\bert{}}\end{sideways}}&identical &	$\textbf{0.38}$&	$0.22$&	$\textbf{0.38}$&	$0.27$&	$0.31$&	$0.29$\\
&synonym	&$0.03$&	$0.08$&	$0.10$&	$0.10$&	$0.06$&	$0.02$\\
&hypernym &	$0.07$&	$0.19$&	$0.14$&	$\textbf{0.40}$&	$0.17$&	$0.16$\\
&hyponym	&$0.17$&	$0.00$&	$0.10$&	$0.15$&	$0.06$&	$0.08$\\
&unrelated/not directly related&	$0.34$&	$\textbf{0.50}$&	$0.29$&	$0.10$&	$\textbf{0.40}$&	$\textbf{0.45}$\\
        \midrule 
\multirow{5}{*}{\begin{sideways}\textbf{\roberta{}}\end{sideways}}&identical &	$\textbf{0.48}$&	$0.22$&	$\textbf{0.52}$&	$0.31$&	$0.27$&	$0.29$\\
&synonym&	$0.00$&	$0.11$&	$0.00$&	$0.10$&	$0.15$&	$0.08$\\
&hypernym&	$0.21$&	$0.19$&	$0.05$&	$\textbf{0.31}$&	$0.08$&	$0.08$\\
&hyponym&	$0.17$&	$0.00$&	$0.10$&	$0.12$&	$0.08$&	$0.08$\\
&unrelated/not directly related&	$0.14$&$\textbf{0.33}$	&	$0.47$&$0.16$&$\textbf{0.47}$	&$\textbf{0.42}$	\\
         \midrule 
\multirow{5}{*}{\begin{sideways}\textbf{\albert{}}\end{sideways}}&identical &$0.28$&$0.14$&$0.29$&$0.20$&$0.15$&$0.14$\\
&synonym&$0.03$&$0.06$&$0.00$&$0.02$&$0.08$&$0.00$\\
&hypernym&$0.21$&$0.19$&$0.24$&$\textbf{0.43}$&$0.10$&$0.16$\\
&hyponym&$0.17$&$0.03$&$0.14$&$0.08$&$0.08$&$0.10$	\\
&unrelated/not directly related&$\textbf{0.31}$&$\textbf{0.58}$&$\textbf{0.33}$&$0.27$&$\textbf{0.58}$&$\textbf{0.59}$\\

\bottomrule
    \end{tabular}
    \caption{Predicted \textbf{types of antecedents} by relativizer in percentage. The type hypernym encompasses also general nouns, determiners, and pronouns.}
    \label{tab:testsem}
\end{table}

\end{document}